\documentclass[journal,10pt]{IEEEtran}

\ifCLASSINFOpdf
\else
\fi
\usepackage[utf8]{inputenc} 
\usepackage[T1]{fontenc}    
\usepackage{hyperref}       
\usepackage{url}            
\usepackage{booktabs}       
\usepackage{amsfonts}       
\usepackage{nicefrac}       
\usepackage{microtype}      
\usepackage{xcolor}         
\usepackage{graphicx}
\usepackage{bm}

\usepackage{amsmath}
\usepackage{amssymb}
\usepackage{mathtools}
\usepackage{amsthm}

\usepackage{bbding}
\usepackage{pifont}
\usepackage{color,xcolor}
\usepackage{wasysym}
\usepackage{mdframed}

\usepackage{color}
\usepackage{subfigure}
\usepackage{booktabs} 
\usepackage{multirow}
\usepackage{multicol}
\usepackage{hyphenat}

\usepackage{float}
\usepackage{subfigure}
\usepackage{makecell}
\usepackage{enumitem}
\usepackage[ruled,boxed,linesnumbered]{algorithm2e}

\usepackage[capitalize,noabbrev]{cleveref}
\usepackage{wrapfig}

\usepackage[textsize=tiny]{todonotes}

\graphicspath{{figs/}}

\def\lyu{\textcolor{black}}

\def\mr{\mathrm}
\def\mb{\mathbf}

\def\mc{\mathcal}

\def\etal{{\em et al.\/}\, }

\def\ie{\textit{i.e.}}
\def\wrt{\textit{w.r.t. }}

\newtheorem{definition}{Definition}
\newtheorem{lemma}{Lemma}
\newtheorem{theorem}{Theorem}
\definecolor{cpink}{RGB}{255,228,196}
\definecolor{cblue}{RGB}{176,196,222}

\begin{document}
%
\title{Elastic Multi-Gradient Descent\\for Parallel Continual Learning}



\author{Fan Lyu$^*$,~\IEEEmembership{Member,~IEEE},
        Wei Feng$^*$,~\IEEEmembership{Member,~IEEE},
        Yuepan Li,
        Qing Sun,\\
        Fanhua Shang,~\IEEEmembership{Member,~IEEE},
        Liang Wan,~\IEEEmembership{Member,~IEEE},
        Liang Wang,~\IEEEmembership{Fellow,~IEEE}
\thanks{F. Lyu and L. Wang is with the Center for Research on Intelligent Perception and Computing (CRIPAC)
        Institute of Automation, Chinese Academy of Sciences (CASIA), Beijing, 100190 China. Contacts:  fan.lyu@cripac.ia.ac.cn, wangliang@nlpr.ia.ac.cn}%
\thanks{W. Feng, Y. Li, Q. Sun, F. Shang, L. Wan are with the College of Intelligence and Computing, Tianjin University, Tianjin, 300350 China. Contacts: (\{yuepanli, sssunqing, fhshang, lwan\}@tju.edu.cn, wfeng@ieee.org)}
\thanks{*Equal Contribution.}
\thanks{This study was done when F. Lyu was working for his PhD in Tianjin University.}
\thanks{The corresponding author is Wei Feng.}
\thanks{Manuscript received ** **, 20**; revised August **, **.}
}

\markboth{Journal of \LaTeX\ Class Files,~Vol.~14, No.~8, August~2015}%
{Shell \MakeLowercase{\textit{et al.}}: Bare Demo of IEEEtran.cls for IEEE Transactions on Magnetics Journals}
%



\IEEEtitleabstractindextext{%
\begin{abstract}
The goal of Continual Learning (CL) is to continuously learn from new data streams and accomplish the corresponding tasks.
Previously studied CL assumes that data are given in sequence nose-to-tail for different tasks, thus indeed belonging to Serial Continual Learning~(SCL).
This paper studies the novel paradigm of Parallel Continual Learning (PCL) in dynamic multi-task scenarios, where a diverse set of tasks is encountered at different time points.
PCL presents challenges due to the training of an unspecified number of tasks with varying learning progress, leading to the difficulty of guaranteeing effective model updates for all encountered tasks.
\lyu{In our previous conference work, we focused on measuring and reducing the discrepancy among gradients in a multi-objective optimization problem, which, however, may still contain negative transfers in every model update.
To address this issue, in the dynamic multi-objective optimization problem, we introduce task-specific elastic factors to adjust the descent direction towards the Pareto front.}
The proposed method, called Elastic Multi-Gradient Descent (EMGD), ensures that each update follows an appropriate Pareto descent direction, minimizing any negative impact on previously learned tasks.
To balance the training between old and new tasks, we also propose a memory editing mechanism guided by the gradient computed using EMGD. This editing process updates the stored data points, reducing interference in the Pareto descent direction from previous tasks.
Experiments on public datasets validate the effectiveness of our EMGD in the PCL setting.



\end{abstract}

\begin{IEEEkeywords}
parallel continual learning, catastrophic forgetting, training conflict, multi-objective optimization, Pareto optimum.
\end{IEEEkeywords}}

\maketitle

\IEEEdisplaynontitleabstractindextext

%
\IEEEpeerreviewmaketitle

\section{Introduction}
\begin{figure}[t]
\centering
\includegraphics[width=\linewidth]{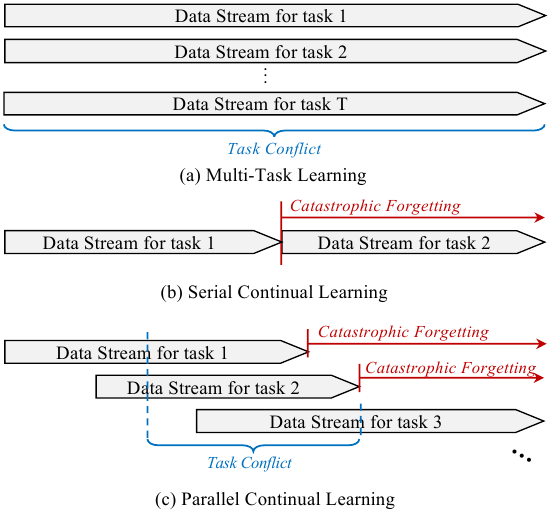}
\vspace{-15px}
\caption{
Comparisons of Multi-Task Learning (MTL), Serial Continual Learning (SCL) and Parallel Continual Learning (PCL).
(a) MTL relies on a fixed number of different tasks without adaptive incremental learning for new tasks.
(b)SCL learns from sequential multiple tasks, where new tasks have to wait for the completion of previous training.
(c) PCL enables training of multiple tasks based on their access time, allowing new tasks to be incorporated immediately.
}
\label{fig:compare}
\end{figure}

%
%
%
%
\IEEEPARstart{H}{ow} to learn a single model for multiple tasks from multiple data sources has been a long-standing research topic~\cite{ma2018modeling, lepikhin2020gshard, barham2022pathways}, with applications in various fields, such as recommendation systems~\cite{zhao2019recommending} and machine translation~\cite{zhang2020share}.
Two major paradigms (\ie, \textit{Multi-Task Learning} (MTL) and \textit{Continual Learning} (CL)) have emerged for learning a unified model, and they differ in terms of their approach to data access. Fig.~\ref{fig:compare}(a) and (b) provide an overview of MTL and CL.

MTL~\cite{Yang2014,Peng2016,Fourure2017} receives a fixed number of data pipes (from one to many) for different tasks, where all tasks are trained in parallel at the same time.
The challenge of MTL is the competition between tasks, \ie, task conflict.
To address the task conflict, traditional MTL solutions can be mainly grouped into feature-based~\cite{maurer2013sparse,wang2015safe} and parameter-based~\cite{thrun1996discovering,barzilai2015convex,jalali2010dirty} approaches.
In recent years, some studies focus on reweighting multi-task training for task balancing, which can be categorized into three kinds, including learning-based methods~\cite{chen2018gradnorm,liu2021towards}, solving-based methods~\cite{sener2018multi}, and calculating-based methods~\cite{liu2019end,chen2020just,yu2020gradient,groenendijk2021multi,lin2021closer}.

Different from MTL, CL~\cite{kirkpatrick2017overcoming,li2017learning,lopez2017gradient} aims to continuously learn new knowledge from a sequence of tasks with non-overlapping data streams over a lifelong time.
Because the data of each task comes in sequence, the challenge of CL is catastrophic forgetting of old tasks or finished tasks.
The existing CL methods are grouped into three types\cite{de2021continual}, \ie, regularization-based methods~\cite{kirkpatrick2017overcoming,chaudhry2018riemannian,dhar2019learning,zenke2017continual,aljundi2018memory,du2022agcn,du2022class}, architecture-based methods~\cite{mallya2018piggyback,yoon2017lifelong,rusu2016progressive,rosenfeld2018incremental}. and rehearsal-based methods~\cite{lopez2017gradient,AGEM,guo2019learning,atkinson2018pseudo,shah2018distillation,pomponi2020pseudo,lyu2021multi}.
Specifically, rehearsal has been proven effective against forgetting, which stores a small sample from old tasks and retrains them together with new tasks.  
In fact, both MTL and CL achieve the goal of a single model for multiple tasks, where the difference between them is the data form as shown in Fig.~\ref{fig:compare}. 



Despite the success of MTL and traditional CL in learning a unified model for multiple tasks, they are still insufficient for learning in a changing environment with multiple data sources. 
On one hand, classical MTL never considers new tasks or environments access in sequence or incrementally, leading to poor adaptation to new knowledge. 
On the other hand, CL can hardly achieve an immediate response to new task demands, new task data streams can only be learned after the previous task training is finished.
We refer to this type of continual learning as \emph{Serial Continual Learning} (SCL).
To address the limitations of MTL and CL, we propose a more general and practical scenario, where new tasks can be accessed and trained at any time, together with any in-training tasks.



In our previous conference work~\cite{lyu2023measuring}, we investigated \emph{Parallel Continual Learning}~(PCL) that cuts off the one-pipe data stream setting in SCL and makes incremental adaption in MTL.
In PCL, a new task demand can arise at any time, allowing for instant response to new task demands without waiting for pending tasks. 
PCL satisfies the requirements of continual learning and immediate response by enabling new tasks to be trained immediately and in parallel with other in-training tasks, as depicted in Fig.\ref{fig:compare}(c). 
PCL faces challenges inherited from both MTL and SCL: \textit{task conflict} and \textit{catastrophic forgetting}. 
The diverse training processes across different tasks can lead to the task conflict issue~\cite{sener2018multi}, in which back-propagation may not guarantee all in-training tasks to converge to their respective optimum. 
Meanwhile, the catastrophic forgetting issue~\cite{french1999catastrophic,kirkpatrick2017overcoming} arises due to the unavailability of finished task data~\cite{Babcock2002}. Notably, these two issues are more complex than those encountered in traditional MTL and SCL, as the access time for new tasks is uncertain and may occur at any time.





\lyu{Nonetheless, our previous study~\cite{lyu2023measuring} only tries to balance multi-tasks in an intuitive way, and ignores the negative transfer in updates, which can not satisfy the Pareto direction. }
In this paper, we address the challenges of catastrophic forgetting and training conflict in PCL by formulating it as a dynamic MTL problem, \lyu{and seek to constrain the update towards the Pareto front.} 
To this end, we propose a simple yet effective approach called Elastic Multi-Gradient Descent (EMGD). EMGD builds on the well-established steepest descent method~\cite{fliege2000steepest} in Multi-Objective Optimization (MOO) and adds an elastic factor for each task.
{The elastic factor is utilized to mitigate the discrepancy between the gradient of each task and the under-optimized gradient, and can be evaluated by gradient magnitude changes and gradient similarities. }
EMGD can be easily solved in its dual form with quadratic programming.
We also prove that EMGD can converge to a Pareto critical point for long-term training.
Additionally, we propose to edit the stored data with the guidance of the Pareto descent direction obtained from EMGD.
By performing such memory editing, the potential task conflict between old and new tasks can be reduced, and the memory can be utilized in subsequent training. 

\lyu{Our previous conference work~\cite{lyu2023measuring} focused on measuring the discrepancy among multiple gradients. We proposed an explicit measurement for the learning from gradient discrepancy in PCL, named asymmetric gradient distance, which considers gradient magnitude ratios and directions, and sets a tolerance for smaller gradients. 
Moreover, an effective optimization strategy was proposed for minimizing the gradient discrepancy to avoid self-interference.
We name the strategy Maximum Discrepancy Optimization (MaxDO), which minimizes the maximum discrepancy from each gradient to the others.
In comparison with our previous work, this work aims to further require a Pareto descent direction achieved from the varied gradients.
To achieve this, in this paper, we transform the PCL task into a dynamic MOO problem, and propose a simple yet effective elastic multi-gradient descent method, which guarantees the training of new tasks while reducing the catastrophic forgetting of previous methods by maintaining the Pareto descent direction in each update.
Compared to the previous work~\cite{lyu2023measuring}, the proposed EMGD method has the following three advantages:
(1) \textit{Pareto descent direction in MOO}. 
This paper is inspired by the classic MGDA and proposes to control the Pareto descent direction, which will never worsen any training task.
(2) \textit{Better rehearsal using Pareto descent direction}. 
This paper proposes to edit the stored memory for better rehearsal, which mitigates the task conflicts within the stored samples, and reduces the catastrophic fgetting.
(3) \textit{Better performance}. 
We show more experimental comparisons with MTL, SCL and our previous work~\cite{lyu2023measuring} and achieve new state-of-the-art results in PCL.}
In our experiments, to evaluate the proposed method, we construct multiple parallel data streams using three popular image classification datasets, where access time for each task varied. 
Our observations show that EMGD can ensure both new task training and old task retention when a new task is introduced.

Our contributions are three-fold:

\begin{enumerate}
    \item We propose PCL, a general scenario to achieve a single model for multiple data sources and tasks. In contrast to MTL and SCL, PCL conducts adaptive training based on the continuously received multiple data streams. At any time, PCL trains an unfixed number of tasks using available data from these streams
    \item We present EMGD, a solution to resolve dynamic task conflicts in PCL, which transforms the PCL task into a dynamic MOO problem. To solve this problem, we introduce task-specific {elastic factors}. EMGD ensures both new task training and old task preservation in PCL. 
    \item We propose a memory editing technique that relies on the target gradient obtained from EMGD, which edits the pixel-level information to mitigate task conflicts within the stored memory, as well as catastrophic forgetting. 
\end{enumerate}


\section{Related Work}

\subsection{Multi-Task Learning}

Multi-Task Learning (MTL)~\cite{Yang2014,Peng2016} is inspired by the human ability to learn multiple tasks simultaneously, addressing sharing information about different problems in the same~\cite{lin2019pareto,sener2018multi,liu2019multi,doersch2017multi} or different~\cite{Tang2020,nam2016learning,tars2018multi,rebuffi2018efficient} domains. In most situations of MTL, the multiple tasks refer to different outputs of the model, such as different classifiers in classification tasks or segmentation and detection tasks in computer vision. 
These tasks may have inter-dependencies that can benefit each other. 
In such scenarios, multi-task learning allows tasks to reuse knowledge from other tasks, reducing the cost of manual annotation.

MTL is based on the fundamental assumption that multiple tasks, when learned together, can mutually reinforce or constrain each other, ultimately leading to the achievement of complex multi-task recognition capabilities~\cite{zhang2021survey}.
In MTL, simultaneously learning multiple tasks can be more challenging. Unlike single-task learning, there may be task conflicts among multiple tasks, manifesting as each task in multi-task learning not performing as well as it would in isolated single-task training. In such cases, improving the performance of one task may lead to a decline in the performance of other tasks, a phenomenon referred to as Negative Transfer~\cite{zhang2020overcoming}. Mitigating the impact of negative transfer and reducing training conflicts are among the primary objectives in multi-task learning.
In summary, traditional MTL solutions can be mainly grouped into three kinds~\cite{zhang2021survey} as follows.

(1) \textit{Feature-based methods} generally assume that multiple tasks are related, with the core focus being on learning shared representations for different tasks~\cite{maurer2013sparse,wang2015safe}. These shared representations encapsulate the shared knowledge among multiple tasks. By leveraging this shared knowledge for each task, it strengthens its representation capabilities, leading to improved performance.
Based on the relationship between the original feature representation and the learned feature representation, feature-based multi-task learning can be further divided into two categories~\cite{zhang2021survey}: feature transformation methods and feature selection methods. Feature transformation methods~\cite{liao2005radial,silver2008inductive,argyriou2008convex,liu2015multi} suggest that shared features among multiple tasks should be transformed through a model, while feature selection methods~\cite{liu2012multi,lee2010adaptive} assume that there is redundancy in the feature information and aim to learn appropriate combinations as new feature representations for all tasks. 

(2) \textit{Parameter-based methods} utilize model parameters from any task to assist in learning the model parameters for other tasks. These methods can be primarily categorized into four approaches~\cite{zhang2021survey}: low-rank methods, task clustering methods, task relationship learning methods, and decomposition methods.
Low-rank methods interpret the correlation between multiple tasks as a low-rank representation of their parameter matrices~\cite{ando2005framework,zhang2005learning}.
Task clustering methods group tasks into clusters based on similarities, where each task cluster contains similar tasks~\cite{thrun1996discovering,yu2005learning}.
Task relationship learning methods aim to automatically learn quantitative relationships between tasks from the data.
Decomposition methods break down the model parameters for all tasks into multiple sub-models, and each sub-model is subject to different regularizations~\cite{bonilla2007multi,chai2009generalization}.
These approaches are designed to leverage the interplay between task parameters to improve the overall performance of MTL.


(3) \textit{Weighting-based approaches} formulate the MTL problem into finding an optimal gradient via weighting sum for updating, and can be categorized into three types.
The learning-based methods~\cite{chen2018gradnorm,liu2021towards} design extra models to learn the appropriate weights for each task.
The solving-based methods~\cite{sener2018multi} formulate the MTL into a multi-objective optimization problem and obtain the Pareto descent direction.
The calculating-based methods~\cite{liu2019end,chen2020just,yu2020gradient,groenendijk2021multi,lin2021closer} project conflict gradients to other spaces to reduce the conflicts.

In this paper, we only consider the multi-source MTL, \ie, each task has its specific data stream without sharing with other tasks.
However, multi-source MTL can only learn from data within a fixed number of domains for a fixed number of tasks, \ie, without the ability to evolve.
Besides, MTL assumes the data are always totally available, which is not realistic for long-term learning.

\subsection{Continual Learning} 

Continual Learning (CL), also known as incremental learning, receives data from new domains in order to satisfy new task demands.
CL imitates the learning way of the human brain, which is manifested as adaptation to the environment. 
In the real world, people encounter new things and acquire new knowledge every day, which requires the human brain to have the ability of continual learning, that is, to keep learning new knowledge while ensuring that the old knowledge is not forgotten.
Similarly, the CL based on neural networks often uses the gradient descent method to update the model. When learning new knowledge, the gradient containing only new knowledge will overwrite the old knowledge within the model parameter, resulting in catastrophic forgetting~\cite{french1999catastrophic}.
According to the incremental carrier of new knowledge, CL can be task-incremental, class-incremental or other forms.
In this paper, we focus on task-incremental and class-incremental CL in our experiments.
The traditional CL has the training tasks in sequence, that is, nose-to-tail.
We name this kind of CL Serial CL (SCL).
The main solutions so far to CL can be roughly categorized into three types~\cite{de2021continual}.

(1)~\textit{Rehearsal}-based methods~\cite{lopez2017gradient,AGEM,guo2019learning,atkinson2018pseudo,shah2018distillation,pomponi2020pseudo}  save previous knowledge, which will be used to remind model together with the current training.
According to different knowledge storage forms, rehearsal-based methods can be divided into~\cite{de2019continual} three categories.
First, rehearsal on raw data, which retrains directly on the stored raw data, and restricts the update of the model parameters~\cite{robins1995catastrophic,lopez2017gradient,AGEM,chaudhry2019on}.
Second, rehearsal on features constrains the parameter update after the feature extracted layer but does not affect the previous neural network layers~\cite{rebuffi2017icarl,chaudhry2021using,buzzega2020dark}.
Third, rehearsal on synthesizing data from generative models, which stores knowledge in the form of model instead of data~\cite{wu2018memory,liu2020generative}.

(2)~\textit{Regularization}-based methods~\cite{kirkpatrick2017overcoming,chaudhry2018riemannian,dhar2019learning,zenke2017continual,aljundi2018memory}  leverage extra regularization terms in the loss function to consolidate previous knowledge when learning from new data.
Kirkpatrick \etal\cite{kirkpatrick2017overcoming} propose elastic weight consolidation, which uses Laplace approximation to estimate the posterior of previous knowledge and uses the Fisher information matrix to constrain the update.
Aljundi \etal\cite{aljundi2018memory} propose another approximation for parameter importance by adding perturbation on model parameters.
Zenke \etal\cite{zenke2017continual} adopt a similar idea, considering the influence of the loss change can be represented by gradient influence.

(3)~\textit{Architecture}-based methods~\cite{mallya2018piggyback,yoon2017lifelong,rosenfeld2018incremental} allocate task-specific parameters or grow new branches for new tasks to bring in new knowledge.
According to the operation on structure, these methods can be divided into expansion and routing.
The expansion-based methods directly expand the model structure, and the new task is trained on the expanded parameters, so as to avoid the damage of the parameters related to the old task~\cite{rusu2016progressive,douillard2022dytox}.
The routing-based methods seek to make different tasks use different parameter paths, which avoid the overlap of important parameters by new tasks~\cite{fernando2017pathnet,serra2018overcoming}.

However, existing CL methods assume that the new task demand appears only after the previous training is done, which is not realistic enough in real-world applications.
In this paper, we consider a more general scenario that the data \wrt~ a new task can access at any time.
That is, different tasks can be trained with data from different domains at the same time, what we call the Parallel Continual Learning~(PCL).

\section{Parallel Continual Learning}

\label{sec:pll}

\subsection{Problem Definition: Parallel Continual Learning}

Suppose we have a sequence of $T$ tasks corresponding to $T$ parallel data streams $\{\mc{D}_1,\cdots, \mc{D}_T\}$, in which each data stream can be accessed and suspended at any time. 
Data streams are composed by non-overlapping sequences of data tuples
$(\mb{x},\mb{y})$, where $\mb{x}$ is an input data point, $\mb{y}$ is a one-hot label and $C_t$ is the class number of task $t$.
For simplicity, we assume that each data stream is stationary, which means that the data are drawn from a fixed, unknown probability distribution. 

We define the start and end time of the $t$-th data stream as $(s_t,e_t)$.
To simplify matters, we assume that $s_{t-1}\le s_t\le 1+\max(e_1,e_2,\cdots,e_{t-1})$ for any task $t>1$.
This means that tasks are accessed in order from $1$ to $T$, and no blank in the timeline where no data stream is active. 
It is worth noting that traditional SCL represents an extreme case of PCL, where all tasks are executed in sequence, \ie, $\forall t>1$, $s_t=e_{t-1}+1$.

To adapt to incremental tasks, a PCL model is composed of a \textit{task-agnostic backbone} with parameter $\bm{\theta}$ and dynamically incremental number of \textit{task-specific classifiers} with parameters $\bm{\theta}_t$.
When a new task is accessed, a new corresponding task-specific classifier is constructed.
The function of both networks is given by $f_{\bm{\theta}}: \mb{x}  \mapsto \mb{v} \in \mathbb{R}^{1 \times D}$ and $f_{\bm{\theta}_t}: \mb{v} \in \mathbb{R}^{1 \times D} \mapsto \mb{p} \in \mathbb{R}^{1 \times C_t}$,
where $\mb{x},\mb{v},\mb{p}$ denote the input data point, the task-agnostic feature and the predicted probability, respectively.

Based on the above concepts, we define Parallel Continual Learning (PCL) as follows.

\begin{definition}[\textbf{Parallel Continual Learning}]
Parallel Continual Learning (PCL) refers to the adaptive training of a model that continuously receives multiple data streams based on their access period. At any time, PCL trains an unfixed number of tasks using available data from these streams.
\end{definition}   

\subsection{Dynamic multi-task learning with {rehearsal}}

\label{sec:dmtl}

To prevent a task from being forgotten when its data stream is finished, we employ the Rehearsal strategy~\cite{lopez2017gradient,AGEM,guo2019learning,atkinson2018pseudo,shah2018distillation,pomponi2020pseudo}, which involves storing a small amount of training data and creating an additional data stream for retraining completed tasks.
Specifically, a portion of the data from each completed task is stored in a fixed set $\mc{D}_0$. At time $t$, PCL training, together with the memory stream $\mc{D}_0$, yields the following formulation for multi-objective empirical risk minimization:
\begin{equation}
\mathop{\min}_{\bm{\theta},\{\bm{\theta}_i|{i\in\mc{T}_t}\} }\quad \left\{\ell_i\left(\mc{D}_i\right)|i\in\mc{T}_t\right\}
\label{eq:pllbase}
\end{equation}
where $\mc{T}_t$ is the task index set with activated data streams at time $t$.
According to the rehearsal strategy, if at least one finished task exists, we include $0$ in $\mc{T}_t$ and use $\ell_0$ to represent the loss for all finished tasks for experience replay.

A naive solution to this dynamic MOO problem is to apply the Average Gradient (AvgGrad), in which the model is updated with the following rules
\begin{align} 
\bm{\theta}\leftarrow\bm{\theta}-\gamma\cdot\frac{1}{|\mc{T}_t|}\sum\nolimits_{i\in\mc{T}_t}\nabla_{\bm{\theta}} \ell_i,\\
\bm{\theta}_i\leftarrow\bm{\theta}_i-\gamma_i\cdot\nabla_{ \bm{\theta_i}} \ell_i,~~\forall i \in\mc{T}_t
\label{eq:gabase}
\end{align}
where $\gamma$ and $\gamma_i$ are the step sizes for optimization.
Since the task-specific classifiers are updated by their own gradients, we focus on the update of the shared backbone $\bm{\theta}$, and let the negative gradient $\mb{g}_i=-\nabla_ {\bm{\theta}} \ell_i$.
However, treating each task equally in the AvgGrad method is naive and does not lead to Pareto optimality~\cite{censor1977pareto}.

\begin{definition}[\textbf{Pareto Optimality}]
~\\
(1) (\emph{Pareto Dominate}) Let $\bm{\theta}_a$, $\bm{\theta}_b$ be two solutions for Problem~\eqref{eq:pllbase}, $\bm{\theta}_a$ is said to dominate $\bm{\theta}_b$ ($\bm{\theta}_a\prec\bm{\theta}_b$) if and only if $\ell_i(\bm{\theta}_a) \le \ell_i(\bm{\theta}_b)$, $\forall i \in \mc{T}$ and $\ell_i(\bm{\theta}_a) < \ell_i(\bm{\theta}_b)$, $\exists i \in \mc{T}$.\\
(2) (\emph{Pareto Critical}) $\bm{\theta}$ is called Pareto critical if no other solution in its neighborhood can have better values in all objective functions.\\
(3) (\emph{Pareto Descent Direction}) If $\bm{\theta}_1$ is not Pareto critical and can be updated to $\bm{\theta}_2$ by gradient $\mb{g}$. If $\bm{\theta}_2\prec\bm{\theta}_1$, say  $\mb{g}$ is a Pareto descent direction.
\label{definition:1}
\end{definition}





An elegant solution for achieving Pareto optimality in MOO is the Steepest Descent Method (SDM)~\cite{fliege2000steepest}. SDM obtains an optimal descent direction $\mb{d}^*$ that satisfies 
\begin{equation}
\label{eq:sgd}
\begin{aligned}
\mathbf{d}^*,\alpha^* = \arg\min_{\mb{d},\alpha}& \quad \alpha+\frac{1}{2}\left\Vert\mb{d}\right\Vert^2,\\
\text{s.t.}&\quad \mb{g}_i^\top \mb{d} \le \alpha,\quad\forall i \in \mc{T}
\end{aligned}
\end{equation}
where the constraints ensure that each task does not conflict with the gradient $\mb{d}$.
Considering the Karush–Kuhn–Tucker (KKT) condition, the dual problem solved by the Multi-Gradient Descent Algorithm (MGDA)~\cite{sener2018multi} is presented as
\begin{equation}
\label{eq:mgda}
\begin{aligned}
\bm{\lambda}^*= \arg\min_{\bm{\lambda}}*\quad  \Big\Vert\sum\nolimits_i \lambda_i \mb{g}_i\Big\Vert^2,\\
\text{s.t.}*\quad \sum\nolimits_i \lambda_i = 1~~\text{and}~~\lambda_i\ge 0, \forall i .
\end{aligned}
\end{equation}
The objective of MGDA is $0$ and the resulting point satisfies the KKT conditions, or the solution gives a Pareto descent direction that improves all tasks~\cite{desideri2012multiple}.

\begin{figure}[t]
\centering
\includegraphics[width=\linewidth]{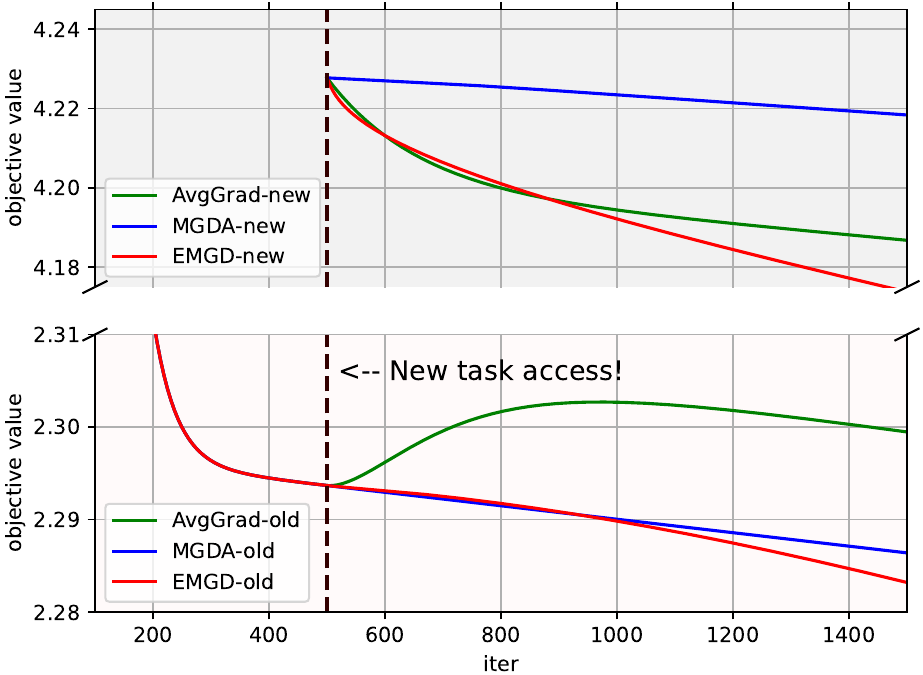}
\vspace{-15px}
\caption{Two synthetic functions are optimized in the PCL setting, where the new task access at iter 500.}
\label{fig:newaccess}
\end{figure}

However, both AvgGrad and MGDA ignore the training variations between any two parallel tasks that may differ a lot.
In Fig.\ref{fig:newaccess}, we conduct a toy PCL experiment with two synthetic functions~\footnote{The two functions are $f_1(x,y) = \log(1+x^2)+0.8\left(1-e^{x}\cdot\sin(y) \right)^2$ and 
$f_2(x,y)= \log(1+y^2)+0.004\left(0.1+e^{y}\cdot\cos(x) \right)^2$. The values of $(x, y)$ are initialized to $(3,3)$.
The optimizer is the steepest gradient descent, and the step size is set to $2e$-$5$.
The total number of iterations is 1,500.
In PCL, we let $f_1(x,y)$ be the old task, and the new task $f_2(x,y)$ is set to access at the 500-th iteration.}.
As we observed, AvgGrad is biased towards the new task with larger gradients, which hinders the optimization of old tasks.
In contrast, MGDA is inclined to update with a conservative Pareto descent direction, which may compromise the new task learning.
To strike a balance among old and new tasks, we propose a novel yet simple Elastic Multi-Gradient Descent (EMGD) method.
EMGD considers the difference in the learning process among tasks and obtains a Pareto descent direction at each iteration that ensures both new task training and old task preservation.

\section{Elastic Multi-Gradient Descent}

\subsection{Elastic constraint for steepest gradient method}

Instead of using direct weighting or solving methods, we seek controllable constraints in PCL. Here, we define an elastic solution for Eq.~\eqref{eq:sgd} with an elastic factor $\sigma > 0$ for $\bm{\theta}$, such that if $\bm{\theta}$ is not Pareto critical, the objective satisfies:
\begin{equation}
\alpha(\bm{\theta}) + \frac{1}{2}||\mb{d}(\bm{\theta})||^2\le\sigma\alpha^*(\bm{\theta})
\end{equation}
where $\alpha^*(\bm{\theta})$ is the optimum of Problem~\eqref{eq:sgd}.
For $\sigma = 1$ only the exact solution satisfies the above inequality.
To further consider the angle between each $\mb{g}_i$ and $\mb{d}$, we generalize the elasticity to each task, and rewrite the Problem~\eqref{eq:sgd} as follows:
\begin{equation}
\label{eq:edga}
\begin{aligned}
\mb{d}^*,\alpha^* = \arg\min_{\mb{d},\alpha}& \quad \alpha+\frac{1}{2}||\mb{d}||^2,\\
\text{s.t.}&\quad \mb{g}_i^\top \mb{d} \le \sigma_i\alpha,\quad\forall i\in\mc{T}.
\end{aligned}
\end{equation}
The only difference of Eq.~\eqref{eq:edga} from SDM is the task-specific elastic factors $\sigma_i\in(0,1]$.
\begin{lemma}
Let $\mb{d}^*(\bm{\theta})$ and $\alpha^*(\bm{\theta})$ be the solution of Eq.~\eqref{eq:edga} under the paramter $\bm{\theta}$. $\mb{d}^*(\bm{\theta})$ and $\alpha^*(\bm{\theta})$ hold the features:\\
(1) if $\bm{\theta}$ is Pareto critical, it has $\mb{d}^*(\bm{\theta})=\mb{0}$ and $\alpha^*(\bm{\theta})=0$;\\    
(2) if $\bm{\theta}$ is not Pareto critical, it has $\alpha^*(\bm{\theta})\le-\frac{1}{2}||\mb{d}^*(\bm{\theta})||^2<0$ and $\mb{g}_i^\top \mb{d}^*(\bm{\theta})\le\sigma_i\alpha^*(\bm{\theta}),~~\forall i$.
\label{lemma:1}
\end{lemma}

\noindent
\textbf{Proof.}
We prove the two features in Lemma~\ref{lemma:1} with the definition of Pareto critical.
Reviewing the definition of Pareto optimality in Definition~\ref{definition:1}

\emph{Feature} 1):
If the solution $\bm{\theta}$ of PCL is Pareto critical, the solution to Problem~\eqref{eq:edga} (EMGD) is $\mb{d}^*(\bm{\theta})=\mb{0}$, then there exists no direction make all objectives get better.
Due to the constraints, we have $\sigma_i\alpha^*(\bm{\theta})\ge\mb{g}_i^\top\mb{d}^*(\bm{\theta}),~\forall i$, thus $\alpha^*(\bm{\theta})=0$.

\emph{Feature} 2):
If the solution $\bm{\theta}$ of PCL is not Pareto critical, it means that $\mb{d}(\bm{\theta})=\mb{0}$ and $\alpha(\bm{\theta})=0$ are the feasible solutions to Problem~\eqref{eq:edga}.
That is, we have
$$
\alpha^*(\bm{\theta}) + \frac{1}{2}\Vert \mb{d}^*(\bm{\theta})\Vert^2 < \alpha(\bm{\theta}) + \frac{1}{2}\Vert \mb{d}(\bm{\theta})\Vert^2 =0.
$$
Thus,
$$
\alpha^*(\bm{\theta}) \le \frac{1}{2}\Vert \mb{d}^*(\bm{\theta})\Vert^2<0.
$$
Moreover, referring to the constraints, we have
$$
\mb{g}_i^\top \mb{d}^*(\bm{\theta})\le\sigma_i\alpha^*(\bm{\theta}),~~\forall i.
$$
All claims of Lemma~\ref{lemma:1} are clear.
$\hfill\blacksquare$

The task-specific elastic factors, which satisfy $0<\sigma<1$, relax the constraint to varying degrees for each task. 
By the Lagrange multipliers for the linear inequality constraints, the dual problem can be rewritten as follows:
\begin{equation}
\begin{aligned}
\bm{\lambda}^*=\min_{ \bm{\lambda}}&\quad  \Big\Vert\sum\nolimits_i \lambda_i \mb{g}_i\Big\Vert^2,\\
\text{s.t.}&\quad  \sum\nolimits_i \lambda_i\sigma_i = 1~~\text{and}~~\lambda_i\ge 0, \forall i.
\end{aligned}
\label{eq:dual}
\end{equation}
The decision space is reduced to the number of in-training tasks, which is significantly smaller than the original problem.
As a quadratic programming problem, this can be easily solved. The main challenge is to determine the appropriate elastic factor $\sigma$ for each task. If so, new tasks with large gradients can be trained quickly, while old tasks or memory tasks with small gradients can be sustained well. 




\begin{figure*}[t]
\centering
\includegraphics[width=.9\linewidth]{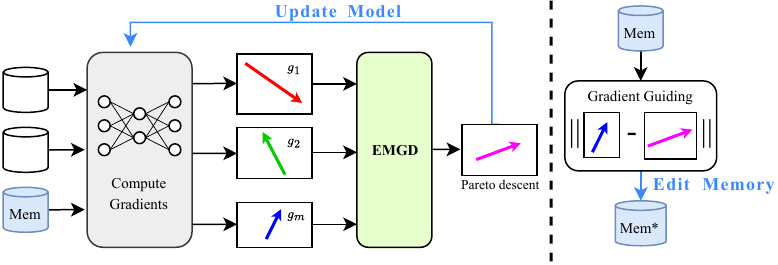}
\caption{
The schematic of the proposed method (one-step).
Together with the memory stream, each activated data stream is put into the model and used to compute the task-specific gradient.
These gradients are sent to EMGD to obtain a Pareto descent direction.
Then, we use the optimal gradient from EMGD to guide the editing on the memory.
}
\label{fig:method}
\end{figure*}

\subsection{Evaluating elastic factor}

Reviewing the {diverse training process} of PCL, we consider the following two goals to set appropriate elastic factors for Problem~\eqref{eq:edga}: 1) \textit{ensuring that new tasks converge as quickly as possible}; 2) \textit{minimizing the impact on old tasks as much as possible}.
To evaluate whether a task satisfies both goals, we propose using two metrics: gradient magnitude change and gradient similarities.


(1) \emph{Gradient Magnitude Change} (GMC): 
The magnitude of the gradient can serve as an indicator to evaluate the degree of convergence or the possibility of being affected by other tasks.
Therefore, we first propose to compute elastic factors by evaluating the gradient norm change for each task as follows:
\begin{equation}
\label{eq:sigma}
\sigma_i =  \cfrac{\exp({m_i})}{\sum_j\exp(m_j)}
\end{equation}
where $m_i$ is the gradient norm with momentum, \ie, $m^t_i = \epsilon_1\cdot m^{t-1}_i +  \epsilon_2\cdot ||\mb{g}^t_i|| $, where $ \epsilon_1$ and $ \epsilon_2$ are empirically set to 0.9 and 0.1, respectively.
The momentum acceleration in mini-batch stochastic gradient descent helps to reduce the instability of the gradient norm.

(2) \emph{{Gradient Similarity}} (GS): 
Gradient similarity between multiple gradients represents the impact of each gradient on other tasks.
Hence, we use the similarity between any two gradients to compute the elastic factors 
\begin{equation}
\label{eq:sigma_2}
\sigma_i =  \frac{\exp(\sum_k\cos(g^t_i, g^t_k))}{\sum_j\exp(\sum_k\cos(g^t_j, g^t_k))}
\end{equation}
Compared to GMC, which only considers the self-change of each gradient, GS takes the relative impact of each gradient on others into account.
We show the details of the full algorithm in Algorithm~\ref{alg:EMGD}.

\begin{algorithm}[t]
\caption{{PCL using EMGD}}
\label{alg:EMGD}
\LinesNumbered
\KwIn{
Random-initialized $\bm{\theta}$, $\bm{\theta}_{1:T}$; Step sizes $\gamma$, $\gamma_{1:T}$ 
}
\KwOut{$\bm{\theta}$, $\bm{\theta}_{1:T}$}

\For{$t$ in timeline}{
$\mc{T}_t\leftarrow$ in-training task index\;
\For{$i\in\mc{T}_t$}{
$\mc{B}_i\sim\mc{D}_i$\;
$\bm{\theta}_i\leftarrow\bm{\theta}_i-\gamma_i \nabla_{\bm{\theta_i}}\ell_i\left(\mc{B}_i\right) $\;
$\mb{g}_i=-\nabla_{\bm{\theta}}\ell\left(\mc{B}_i\right)$\;
}
$\mc{B}_\mr{m}\sim\mc{M}$\;
$\{\sigma_i|\forall i \in \mc{T}_t\}\leftarrow$ Compute elastic factors using GMC (Eq.~\eqref{eq:sigma}) or GS (Eq.~\eqref{eq:sigma_2})\;
$\bm{\lambda}^*\leftarrow$ Solving Problem~\eqref{eq:dual} with all $\sigma$\;
$\bm{\theta}\leftarrow\bm{\theta}+\gamma\cdot\sum_i\lambda_i\mb{g}_i$\;
{
\For{$\mb{x}\in\mc{B}_\mr{m}$}{
$\mb{x}\leftarrow\mb{x}-\alpha \nabla_{\mb{x}}\left\| \mb{g}(\mb{x})-\sum_i\lambda_i\mb{g}_i \right\|^2$
} }
} 
\end{algorithm}

\section{{Gradient-guided Memory Editing}}

As mentioned above, our proposed method, EMGD, aims to balance the training of old and new tasks by taking the difference in gradient magnitudes or similarities into account.
Notably, the issue of catastrophic forgetting in rehearsal is rooted in the quality of the stored memory.
In the case of SCL, existing rehearsal methods are typically based on the assumption that the stored data points are used solely to represent the current task. However, this approach may result in a diminishing utility of the stored memory over time.
To overcome this limitation, the Gradient-based Memory EDiting (GMED)~\cite{GMED} method focuses on refreshing the memory by evaluating the interference between old and new tasks. It achieves this by evaluating loss difference on the same memory data point $\mathbf{x}$ before and after training on the new task. 

Given a model with parameter $\bm{\theta}$, GMED needs to first store the model $\bm{\theta}'$ updated with a target gradient $\mb{d}$.
To reduce the interference, one can edit the input data $\mb{x}$ as follows:
\begin{equation}
\mb{x}=\mb{x}-\alpha \nabla_{\mb{x}}\left\| \ell(\mb{x},\bm{\theta})-\ell(\mb{x},\bm{\theta}') \right\|^2
\end{equation}
where $\bm{\theta}' = \bm{\theta} - \alpha \mb{d}$.
In this way, it is possible that the difference in loss values between the old and new tasks is negligible, but the update gradients point in completely different directions. This discrepancy makes it inaccurate to edit the memory based solely on the loss increase.

\begin{figure*}[t]
\centering
\includegraphics[width=.95\linewidth]{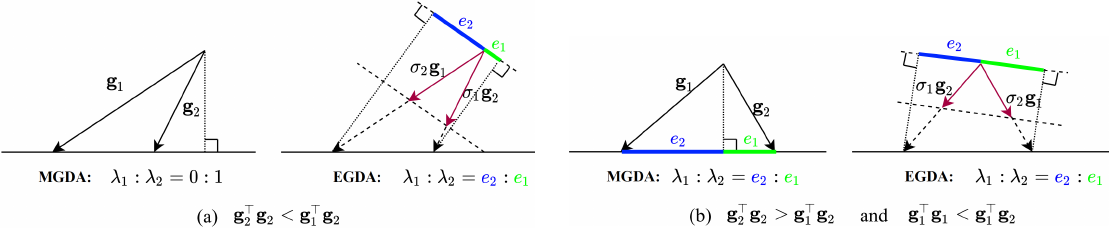}
\caption{Compared to MGDA~\cite{sener2018multi}, EMGD can control the ratio of $\lambda_2$ and $\lambda_1$ in two task scenario. The ratio highly depends on the value of $\sigma$.}
\label{fig:ratio}
\end{figure*}


However, in PCL, simply refreshing the memory based on the \textit{loss difference} is not sufficient. Since PCL utilizes parallel data streams, it is important to accurately edit the memory to minimize both forgetting and conflicts between old and new tasks. To achieve this, we propose to edit the memory based on the \textit{gradient difference}, which can be computed using the optimal gradient $\mb{d}$ as a reference:
\begin{equation}
\mb{x}=\mb{x}-\alpha \nabla_{\mb{x}}\left\| \mb{g}(\mb{x})-\mb{d} \right\|^2
\label{eq:me}
\end{equation}
where 
\begin{equation}
  \nabla_{\mb{x}}\left\| \mb{g}(\mb{x})-\mb{d} \right\|^2 = 2\left( \frac{\partial \ell (\mb{x},\bm{\theta})}{\partial\bm{\theta}}-\mb{d}\right)^\top\frac{\partial^2 \ell (\mb{x},\bm{\theta})}{\partial\bm{\theta}\partial\mb{x}}.
\end{equation}
As shown in Algorithm~\ref{alg:EMGD} (L12-14) and Fig.~\ref{fig:method}, the proposed memory editing based on gradient guidance is easy to implement with a second-order derivate.

\subsection{Convergence analysis}
\label{sec:converge}

In this subsection, we prove that our method can converge to a Pareto critical point.

\begin{theorem}
The accumulation point of the continuous optimization by Algorithm~\ref{alg:EMGD} is Pareto Critical.
\end{theorem}

\noindent
\textbf{Proof.}
Let the solution at the $k$-th iteration be $\bm{\theta}_k$, and the corresponding objective of Problem~\eqref{eq:edga} is $\alpha(\bm{\theta}_k)+\frac{1}{2}||\mb{d}(\bm{\theta}_k)||^2$.
Suppose an accumulation point is $\hat{\bm{\theta}}$ and
\begin{equation}
\lim_{k\rightarrow \infty}\ell(\bm{\theta}_k)=\ell(\hat{\bm{\theta}}).
\end{equation}
Thus,
\begin{equation}
\lim_{k\rightarrow \infty}|\ell(\bm{\theta}_{k+1})-\ell(\bm{\theta}_k)|=0.
\end{equation}
For each task, according to the Taylor Expansion, we have
\begin{equation}
\lim_{k\rightarrow \infty}\ell_i(\bm{\theta}_{k+1})-\ell_i(\bm{\theta}_k)\le \gamma\cdot \mb{g}_i^\top\mb{d}(\bm{\theta}_k)\le 0,\quad \forall i
\end{equation}
where $\gamma$ is the step size for optimization.
Since $\ell_i(\bm{\theta}_{k+1})-\ell_i(\bm{\theta}_k)$ serves as the lower bound of $\gamma\cdot \mb{g}_i^\top \mb{d}$, we have
\begin{equation}
\lim_{k\rightarrow \infty} \mb{g}_i^\top \mb{d}(\bm{\theta}_k)=0,\quad \forall i.
\end{equation}
According to Lemma~\ref{lemma:1}, we have $
\mb{g}_i^\top \mb{d}(\bm{\theta}_k) \le\sigma_i{\alpha}(\bm{\theta}_k)$ and $\alpha(\bm{\theta}_k)\le-\frac{1}{2}||\mb{d}(\bm{\theta}_k)||^2$, thus
\begin{equation}
\lim_{k\rightarrow \infty}{\alpha}(\bm{\theta}_k)=0,\quad \lim_{k\rightarrow \infty}\mb{d}(\bm{\theta}_k)=\mb{0}.
\end{equation}
Since $\hat{\bm{\theta}}$ is the accumulated point of $\bm{\theta}^k$, $\hat{\bm{\theta}}$ is Pareto critical.
$\hfill\blacksquare$






\subsection{Discussion}

\subsubsection{The limitation of MGDA}

Suppose we have two gradients $\mb{g}_1$ and $\mb{g}_2$ w.r.t. two tasks 1 and 2, and $||\mb{g}_1||>||\mb{g}_2||$.
We build the objective of MGDA
\begin{equation}
\label{eq:mgda2}
\begin{aligned}
\lambda_1^*,\lambda_2^*= \min_{\lambda_1,\lambda_2}\quad  &\big\Vert\ \lambda_1 \mb{g}_1+\lambda_2\mb{g}_2\big\Vert^2,\\
\text{s.t.}\quad  &\lambda_1+\lambda_2 = 1~~\text{and}~~\lambda_i\ge 0, \forall i \in\{1,2\}.
\end{aligned}
\end{equation}
It is easy to obtain the optimal $\lambda_1^*,\lambda_2^*$ by solving the above problem, and we have
\begin{equation}
\label{eq:mgda2solution}
\lambda_1^* = \max\left(\min\left(\cfrac{\mb{g}_2^\top\mb{g}_2-\mb{g}_1^\top\mb{g}_2}{\left\Vert\mb{g}_1-\mb{g}_2\right\Vert^2},1 \right),0\right),\quad
\lambda_2^* = 1-\lambda_1^*.
\end{equation}
It is easy to  know that Eq.~\eqref{eq:mgda2solution} can be piecewise as
\begin{equation}
\label{eq:mgda3}
\lambda_1^*,\lambda_2^* = \left\{
\begin{aligned}
&1,0,~~~~~\text{if}~~ \mb{g}_1^\top\mb{g}_2\ge\mb{g}_1^\top\mb{g}_1\\
&0,1,~~~~~\text{if}~~ \mb{g}_1^\top\mb{g}_2\ge\mb{g}_2^\top\mb{g}_2\\
&\frac{\mb{g}_2^\top\mb{g}_2-\mb{g}_1^\top\mb{g}_2}{\left\Vert\mb{g}_1-\mb{g}_2\right\Vert^2}, \frac{\mb{g}_1^\top\mb{g}_1-\mb{g}_1^\top\mb{g}_2}{\left\Vert\mb{g}_1-\mb{g}_2\right\Vert^2},~~~ \text{otherwise}.
\end{aligned}
\right.
\end{equation}
Since $||\mb{g}_1||>||\mb{g}_2||$, we have $\mb{g}_1^\top\mb{g}_1>\mb{g}_2^\top\mb{g}_2$ and $\mb{g}_1^\top\mb{g}_1>\mb{g}_1^\top\mb{g}_2$.

If $\mb{g}_1^\top\mb{g}_1>\mb{g}_1^\top\mb{g}_2\ge\mb{g}_2^\top\mb{g}_2$, we have $\lambda_1^*=0,\lambda_2^*=1$ that $\lambda_1^*<\lambda_2^*$.

If $\mb{g}_1^\top\mb{g}_1>\mb{g}_2^\top\mb{g}_2\ge\mb{g}_1^\top\mb{g}_2$, we have $\lambda_1^*=\cfrac{\mb{g}_2^\top\mb{g}_2-\mb{g}_1^\top\mb{g}_2}{\left\Vert\mb{g}_1-\mb{g}_2\right\Vert^2}, \lambda_2^*=\cfrac{\mb{g}_1^\top\mb{g}_1-\mb{g}_1^\top\mb{g}_2}{\left\Vert\mb{g}_1-\mb{g}_2\right\Vert^2}$ that $\lambda_1^*<\lambda_2^*$.

To sum up, if we have $||\mb{g}_1||>||\mb{g}_2||$ in MGDA, then $\lambda_1<\lambda_2$ and the update $\mb{d}= -\lambda_1\mb{g}_1-\lambda_2\mb{g}_2$ has preference towards the small gradient.
And this preference depends on the gradient norm difference.
A similar phenomenon can be promoted to multiple tasks.
$\hfill\blacksquare$

\begin{table*}[t]
\centering
\caption{
Task-incremental PCL final performance comparisons (avg $\pm$ std). 
}
\resizebox{\linewidth}{!}{
\begin{tabular}{l|c|rr|rr|rr}
\toprule
\textbf{Method} & \multirow{2}{*}{\textbf{Type}} &  \multicolumn{2}{c|}{\textbf{PS-EMNIST}} &  \multicolumn{2}{c|}{\textbf{PS-CIFAR-100}} &  \multicolumn{2}{c}{\textbf{PS-ImageNet-TINY}} \\
(Rehearsal) & & \multicolumn{1}{c}{$A_{\bar{e}}$ (\%)} & \multicolumn{1}{c|}{$F_{\bar{e}}$ (\%)} & \multicolumn{1}{c}{$A_{\bar{e}}$ (\%)} & \multicolumn{1}{c|}{$F_{\bar{e}}$ (\%)}& \multicolumn{1}{c}{$A_{\bar{e}}$ (\%)} & \multicolumn{1}{c}{$F_{\bar{e}}$ (\%)}  \\
\midrule
AvgGrad	& MTL  & $89.344\pm0.231$ & $-5.287\pm0.216$ & $47.579\pm0.089$ & $23.691\pm0.432$ & $38.392\pm0.076$ & $0.764\pm0.139$\\
MGDA~\cite{desideri2012multiple} & MTL & $84.887\pm0.469$ & $-4.301\pm0.818$ & $48.957\pm0.451$ & $25.088\pm0.203$  & $38.058\pm0.740$ & $1.465\pm0.490$ \\
GradNorm~\cite{chen2018gradnorm} & MTL   & $87.888\pm0.158$ & $-6.197\pm0.183$& $47.210\pm1.323$ & $23.474\pm0.688$& $38.226\pm0.769$ & $1.647\pm0.667$ \\
DWA~\cite{liu2019end}   & MTL      & $88.405\pm0.322$ & $-5.452\pm0.338$ & $44.969\pm0.378$ & $22.682\pm0.537$ & $34.290\pm1.099$ & $-1.053\pm0.866$ \\
PCGrad~\cite{yu2020gradient} & MTL  & $89.698\pm0.164$ & $-4.921\pm0.121$ &  $47.026\pm0.538$ & $23.244\pm0.740$ & $39.427\pm1.275$ & $2.017\pm0.769$ \\
RLW~\cite{lin2021closer} 	      & MTL      & $89.288\pm0.218$ & $-5.226\pm0.204$ & ${47.574}\pm0.349$ & $23.833\pm0.117$ & $38.531\pm1.610$ & $1.332\pm1.148$ \\
\midrule
AGEM~\cite{AGEM} & SCL	 & $87.022\pm0.519$ & $-7.646\pm0.483$  & $27.379\pm0.585$ & $5.416\pm0.851$ & $28.530\pm0.994$ & $-7.070\pm1.410$ \\
GMED~\cite{GMED} & SCL	 & $85.471\pm0.324$ & $-8.875\pm0.335$  & $49.094\pm1.792$&$18.356\pm1.345$& $34.495\pm1.568$ & $-0.640\pm1.799$  \\
ER~\cite{chaudhry2019on} & SCL	 & $89.106\pm0.315$ & $-5.525\pm0.207$ & $47.324\pm0.584$ & $23.330\pm0.762$ & $35.950\pm0.763$ & $-0.767\pm0.591$ \\
MEGA~\cite{guo2019learning} & SCL	 & $86.262\pm0.582$ & $-4.753\pm0.295$ & ${47.745}\pm0.460$ & $23.718\pm0.832$ & $38.386\pm1.234$ & $2.655\pm0.824$\\
MDMTR~\cite{lyu2021multi} & SCL	 & $73.115\pm0.382$ & $-16.359\pm0.178$ & $38.865\pm1.691$  & $18.297\pm1.052$ &$23.422\pm0.273$ & $-1.544\pm0.389$\\
DER++~\cite{buzzega2020dark} & SCL	 & $89.654\pm0.451$ & $-4.979\pm0.293$  &$49.276\pm0.480$&$22.634\pm0.165$& $40.039\pm0.039$ & $2.601\pm0.643$\\
\midrule
MaxDO~\cite{lyu2023measuring} & PCL	 & $90.189\pm0.314$ & $-4.258\pm0.311$  &$50.203\pm0.978$&$24.510\pm0.092$& $40.770\pm0.354$ & $3.119\pm0.450$\\
EMGD(GMC) & PCL	 & $90.442\pm0.144$ & $-\textbf{4.142}\pm0.182$ & $52.798\pm0.768$ & $\textbf{26.858}\pm0.426$ & $46.666\pm0.095$ & $8.934\pm0.350$\\
EMGD(GMC)+ME & PCL	 & $90.402\pm0.102$ & $-4.217\pm0.093$ & $\textbf{56.611}\pm0.102$ & ${23.970}\pm0.424$ & $48.746\pm0.413$ & $6.086\pm0.499$\\
EMGD(GS)     & PCL   & $90.348\pm0.157$ & $-4.265\pm0.187$ & $52.389\pm1.109$ & $26.530\pm0.524$ & $46.213\pm0.334$ & $7.830\pm0.715$\\
EMGD(GS)+ME        & PCL	 & $\textbf{90.524}\pm0.038$ & $-{4.501}\pm0.060$ & $56.378\pm0.244$ & $22.760\pm0.399$  & $\textbf{49.269}\pm0.285$ & $\textbf{7.744}\pm0.481$ \\
\bottomrule
\end{tabular}}
\label{tab:main_task}
\end{table*}

\begin{table*}[t]
\centering
\caption{
Class-incremental PCL final performance comparisons (avg $\pm$ std). 
}
\resizebox{\linewidth}{!}{
\begin{tabular}{l|c|rr|rr|rr}
\toprule
\textbf{Method} & \multirow{2}{*}{\textbf{Type}} &  \multicolumn{2}{c|}{\textbf{PS-EMNIST}} &  \multicolumn{2}{c|}{\textbf{PS-CIFAR-100}} &  \multicolumn{2}{c}{\textbf{PS-ImageNet-TINY}} \\
(Rehearsal) & & \multicolumn{1}{c}{$A_{\bar{e}}$ (\%)} & \multicolumn{1}{c|}{$F_{\bar{e}}$ (\%)} & \multicolumn{1}{c}{$A_{\bar{e}}$ (\%)} & \multicolumn{1}{c|}{$F_{\bar{e}}$ (\%)}& \multicolumn{1}{c}{$A_{\bar{e}}$ (\%)} & \multicolumn{1}{c}{$F_{\bar{e}}$ (\%)}  \\
\midrule
AvgGrad	& MTL                            & $43.823\pm0.566$ & $-44.484\pm0.596$ & $8.921\pm0.276$ & $-4.985\pm0.702$ & $7.511\pm0.464$ & $-15.825\pm0.228$\\
MGDA~\cite{desideri2012multiple} & MTL & $52.823\pm0.201$ & $-7.753\pm0.661$  & $11.323\pm0.282$ & $-4.065\pm0.725$ & $8.318\pm0.155$ & $-13.118\pm0.739$\\
GradNorm~\cite{chen2018gradnorm} & MTL                         & $43.292\pm0.590$ & $-44.824\pm0.688$& $9.477\pm0.222$ & $-4.403\pm1.236$ & $6.395\pm0.130$ & $-15.869\pm0.336$ \\
DWA~\cite{liu2019end}   & MTL      & $42.734\pm0.211$ & $-38.944\pm0.581$ & $3.519\pm0.116$ & $-7.470\pm0.244$ & $3.003\pm0.139$ & $-12.774\pm0.158$\\
PCGrad~\cite{yu2020gradient} & MTL  & $45.035\pm0.273$ & $-43.309\pm0.176$ & $10.140\pm0.253$ & $-5.149\pm0.553$ & $7.789\pm0.237$ & $-15.860\pm0.318$\\
RLW~\cite{lin2021closer} 	      & MTL      & ${44.595}\pm0.480$ & $-43.424\pm0.325$ & $9.957\pm0.135$ & $-5.173\pm0.613$ &$7.419\pm0.258$&$-16.079\pm0.255$ \\
\midrule
AGEM~\cite{AGEM} & SCL	 & $26.249\pm0.511$ & $-62.480\pm0.527$ & $3.374\pm0.098$ & $-10.462\pm0.543$ & $3.379\pm0.149$ & $-18.514\pm0.516$\\
GMED~\cite{GMED} & SCL	 & $22.694\pm0.153$&$-65.805\pm0.255$& $4.941\pm0.407$ & $-12.837\pm0.710$  & $3.386\pm0.348$ & $-19.060\pm0.682$ \\
ER~\cite{chaudhry2019on} & SCL	  & $43.474\pm0.860$ & $-44.597\pm0.861$ & $9.317\pm0.339$ & $-5.337\pm0.626$ & $6.126\pm0.277$ & $-13.686\pm0.290$\\
MEGA~\cite{guo2019learning} & SCL	 & ${51.797}\pm0.607$ & $-5.358\pm0.865$ & $12.245\pm0.388$ & $0.380\pm0.503$ & $10.082\pm0.302$ & $-2.375\pm0.119$ \\
MDMTR~\cite{lyu2021multi} & SCL	  & $42.199\pm1.100$  & $-32.140\pm0.996$ &$5.525\pm0.260$ & $-3.454\pm0.145$ & $2.793\pm0.042$ & $-12.992\pm0.082$\\
DER++~\cite{buzzega2020dark} & SCL	 & $50.039\pm0.520$ & $-43.422\pm0.439$  &$11.022\pm0.642$&$-1.653\pm0.120$&$9.190\pm0.429$&$-14.049\pm0.227$\\
\midrule
MaxDO~\cite{lyu2023measuring} & PCL	 & $53.139\pm0.156$ & $-11.903\pm0.476$  &$12.237\pm0.176$&$-2.280\pm0.270$& $9.532\pm0.363$ & $-12.511\pm0.610$\\
EMGD(GMC) & PCL	 & $53.441\pm0.341$ & $-13.250\pm0.904$ & $13.705\pm0.633$ & $0.367\pm1.299$ & $11.512\pm0.216$ & $-11.748\pm0.331$\\
EMGD(GMC)+ME & PCL	  & $\textbf{56.891}\pm0.062$ & $\textbf{-18.515}\pm0.211$ & $\textbf{16.362}\pm0.294$ & $\textbf{-1.654}\pm0.808$ & $12.641\pm 0.921$ & $-9.890\pm0.743$\\
EMGD(GS)     & PCL  & $53.425\pm0.279$ & $-25.981\pm0.587$ & $13.601\pm0.374$ & $1.530\pm0.524$ & $11.588\pm0.508$ & $-11.324\pm1.072$\\
EMGD(GS)+ME        & PCL & $56.031\pm0.254$&$-17.826\pm0.296$& $16.189\pm1.343$ & $-2.937\pm1.490$  & $\textbf{12.796}\pm0.398$ & $\textbf{-9.093}\pm0.831$ \\
\bottomrule
\end{tabular}}
\label{tab:main_class}
\end{table*}

\begin{figure*}[t]
\centering
\subfigure[EMGD]{\centering
  \includegraphics[width=0.495\textwidth]{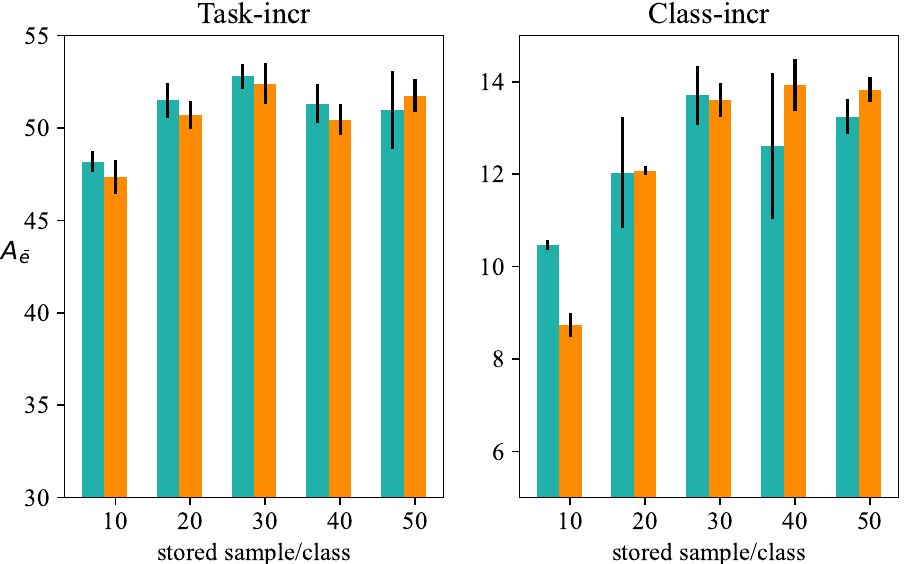}}
\subfigure[EMGD+ME]{\centering
\includegraphics[width=0.495\textwidth]{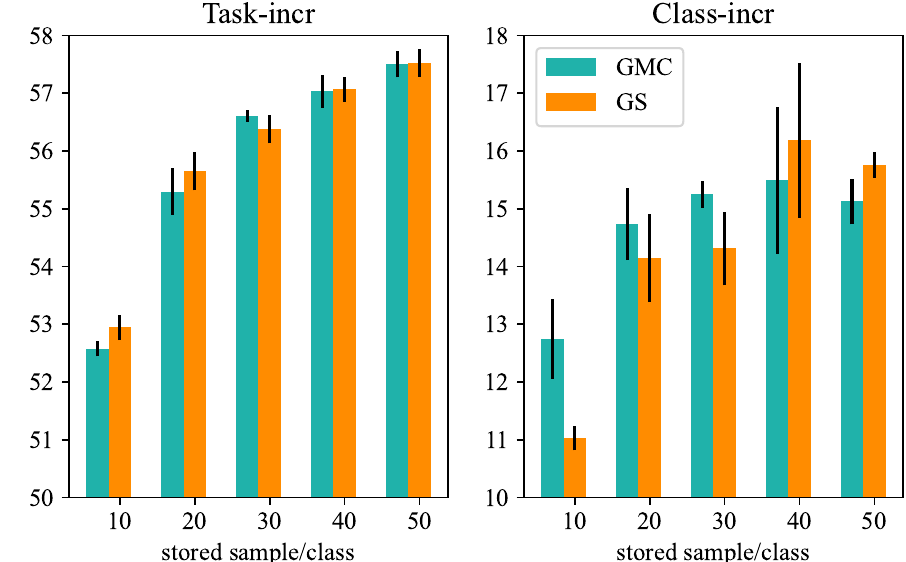}}
\caption{Performance comparisons with different memory size on PS-CIFAR-100.}
\label{fig:var_mem_2}
\end{figure*}

\subsubsection{Comparing MGDA with EMGD}
\label{sec:discuss}

In this subsection, we illustrate why EMGD can control the Pareto descent direction against MGDA.
Let us consider the situation of two tasks at some steps, two gradients $\mb{g}_1$ and $\mb{g}_2$ are obtained and $||\mb{g}_1||>||\mb{g}_2||$.
For MGDA, by solving Eq.~\eqref{eq:mgda}, we have $\lambda_1<\lambda_2$, which means that MGDA prefers to keep the objective of tasks with a small gradient norm.


In the scenario with only two tasks, the objective of EMGD is represented by
\begin{equation}
\label{eq:9}
\begin{aligned}
\lambda_1^*,\lambda_2^*=&\arg\min_{\lambda_1,\lambda_2} \quad\big\Vert\lambda_1 \mb{g}_1+\lambda_2\mb{g}_2\big\Vert^2,\\
&\text{s.t.}\quad  \lambda_1\sigma_1 + \lambda_2\sigma_2 = 1,\lambda_1\ge 0,~~\lambda_2\ge 0.
\end{aligned}
\end{equation}
{If $\sigma_1\mb{g}_2^\top\mb{g}_2<\sigma_2\mb{g}_1^\top\mb{g}_2$, we have $ \lambda_1^*=0$ and $ \lambda_2^*=\frac{1}{\sigma_2}$.
If $\sigma_2\mb{g}_1^\top\mb{g}_1<\sigma_1\mb{g}_1^\top\mb{g}_2$, we have $ \lambda_1^*=\frac{1}{\sigma_1}$ and $ \lambda_2^*=0$.}
Otherwise, we have
\begin{equation}
\lambda_1^*= \frac{\sigma_1\mb{g}_2^\top\mb{g}_2-\sigma_2\mb{g}_1^\top\mb{g}_2}{\big\Vert \sigma_2\mb{g}_1-\sigma_1\mb{g}_2\big\Vert^2}, \quad \lambda_2^*=\frac{\sigma_2\mb{g}_1^\top\mb{g}_1-\sigma_1\mb{g}_1^\top\mb{g}_2}{\big\Vert \sigma_2\mb{g}_1-\sigma_1\mb{g}_2\big\Vert^2}.
\end{equation}
In this situation, 
\begin{equation}
\frac{\lambda_1^*}{\lambda_2^*}
=
\frac{\sigma_1\mb{g}_2^\top\mb{g}_2-\sigma_2\mb{g}_1^\top\mb{g}_2}{\sigma_2\mb{g}_1^\top\mb{g}_1-\sigma_1\mb{g}_1^\top\mb{g}_2}
=
\frac{||\mb{g}_2||\cos\langle \sigma_2\mb{g}_1-\sigma_1\mb{g}_2, \mb{g}_2\rangle}{||\mb{g}_1||\cos\langle \sigma_1\mb{g}_2-\sigma_2\mb{g}_1, \mb{g}_1\rangle}.
\end{equation}
As shown in Fig.~\ref{fig:ratio}, the relative ratio between $\lambda_1^*$ and $\lambda_2^*$ is equivalent to the ratio of gradient projection from two tasks ($\mb{g}_1$ and $\mb{g}_2$) to $\sigma_2\mb{g}_1-\sigma_1\mb{g}_2$.
Note that in MGDA, $\sigma_1=\sigma_2=1$, the ratio depends highly on the norm of gradient.
Instead, EMGD adjusts the ratio via elastic factors.

\section{Experiment}

\label{sec:ex}

\subsection{Dataset construction}

In our experiments, we conduct transformations on three image recognition datasets to suit the requirements of PCL. To ensure effective transformation, we need to fulfill several requirements:
(1) Random label sets for each task: Each data stream had a different label space, with randomly assigned labels for each task;
(2) Random timeline for each label set: The debut of each task could occur at any time between the first access of the previous and subsequent tasks. 
To simplify the process, we omitted any periods where no data streams were present.
The three datasets are as follows:

\begin{itemize}
	\item \textit{Parallel Split EMNIST} (PS-EMNIST): We split EMNIST (62 classes) into 5 tasks and randomly generate 3 label sets for each task and 3 timelines for each label set (say 9 different situations). The size of the label set for each task, i.e., the number of classes, is set to larger than 2 while no more than 15.
	\item \textit{Parallel Split CIFAR-100} (PS-CIFAR-100): We split CIFAR-100 into 20 tasks and randomly generate 3 label sets for each task and 3 timelines for each label set. The size of the label set for each task is set to larger than 2 while no more than 15.
	\item \textit{Parallel Split ImageNet-TINY} (PS-ImageNet-TINY): We split it into 20 tasks w.r.t. random 3 label sets, and each label set has 3 randomly generated timelines. The size of the label set for each task is set to larger than 5 while no more than 20.
\end{itemize}

Note that for each task, we randomly generated 3 label sets and 3 timelines. To ensure fairness in our comparisons, we conducted each experiment using 3 fixed seeds ranging from 1,234 to 1,236. This resulted in a total of 27 different experiments for each dataset.
In our experiments, each subset of data was used only once and then discarded. We considered two types of incremental scenarios in continual learning: task-incremental and class-incremental. Task-incremental learning involved having task IDs available during both the training and testing phases. On the other hand, class-incremental learning did not include task IDs, making it a more challenging scenario.

\subsection{Experiment details}

The input batch size is set to 128 for all three datasets.
We take a 2-layer MLP as the backbone network for PS-EMNIST, and a Resnet-18~\cite{he2016deep} for PS-CIFAR-100 and PS-ImageNet-Tiny, respectively.
In the rehearsal strategy, we store 30 samples per class for PS-CIFAR-100, 5 for PS-EMNIST and 15 for PS-ImageNet-Tiny. 


To evaluate PCL, we compute the average accuracy and the forgetting following previous continual learning studies such as~\cite{lopez2017gradient,AGEM,GMED}.
Suppose $\bar{e}=\max(e_1,e_2,\cdots,e_T)$, the two metrics are computed as follows:
\begin{align}	
A_{\bar{e}}=\frac{1}{T} \sum\nolimits_{t=1}^T a_{\bar{e}}^t,
,\quad
F_{\bar{e}}=\frac{1}{T} \sum\nolimits_{t=1}^T a^t_{\bar{e}}-a_{e_t}^t
\end{align}
where $a^j_k$ is the mean testing accuracy of task $j$ at time $k$.
The $A_{\bar{e}}$ denotes the final average accuracy on all the tasks, and the $F_{\bar{e}}$ (also known as backward transfer) means the final performance drop compared to each task was first trained.



\begin{figure*}[t]
\centering
\subfigure[PS-EMNIST]{\centering
  \includegraphics[width=\linewidth]{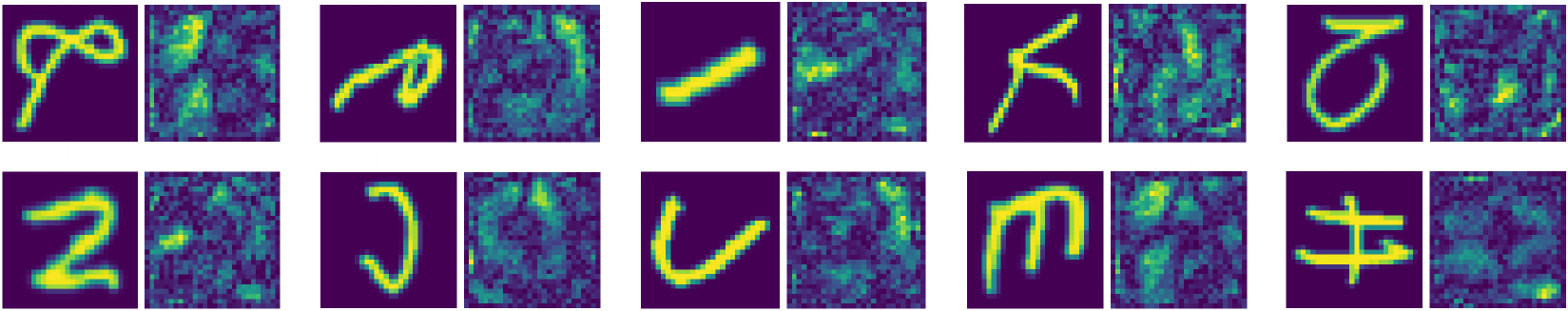}}
\subfigure[PS-CIFAR-100]{\centering
  \includegraphics[width=\linewidth]{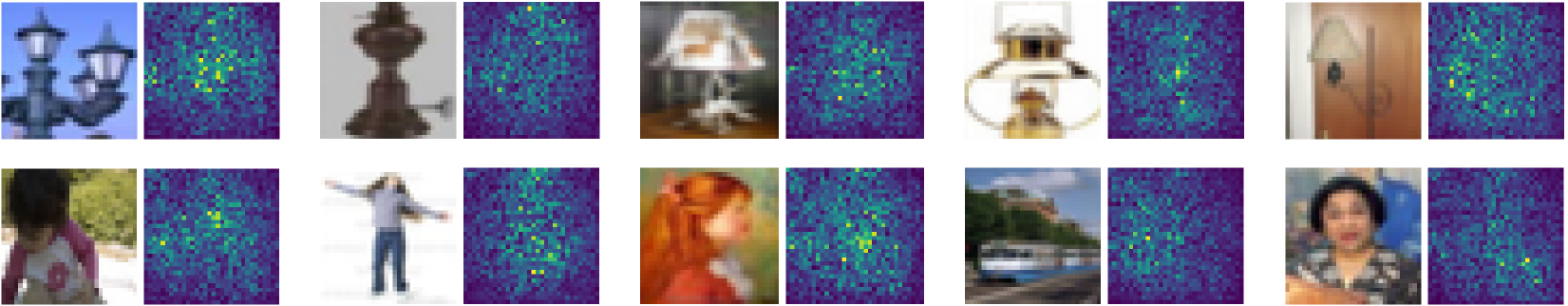}}
\subfigure[PS-ImageNet-TINY]{\centering
  \includegraphics[width=\linewidth]{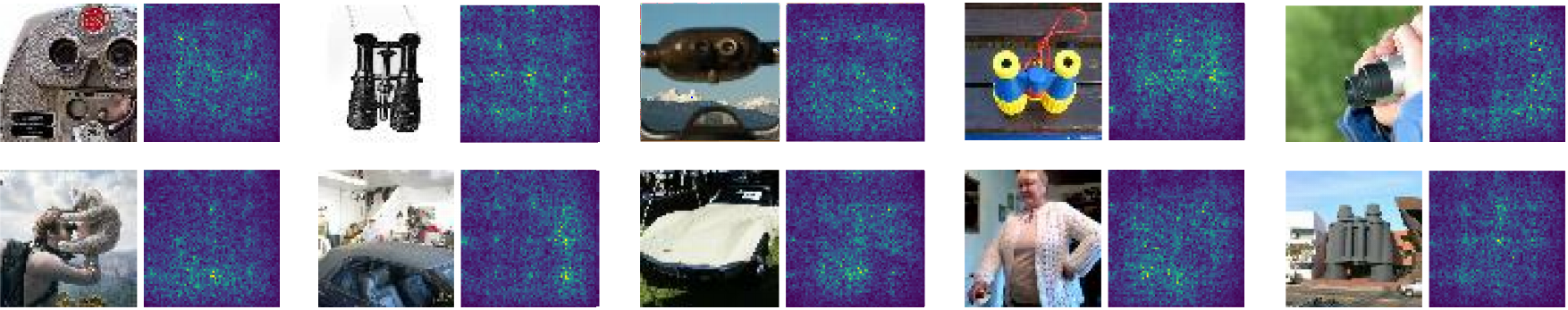}}
\caption{Visualization of editing gradients. Left: Stored images, Right: Normalized gradient.}
\label{fig:vis}
\end{figure*} 

\begin{figure}[t]
\centering
\subfigure[PS-EMNIST]{\centering
  \includegraphics[width=\linewidth]{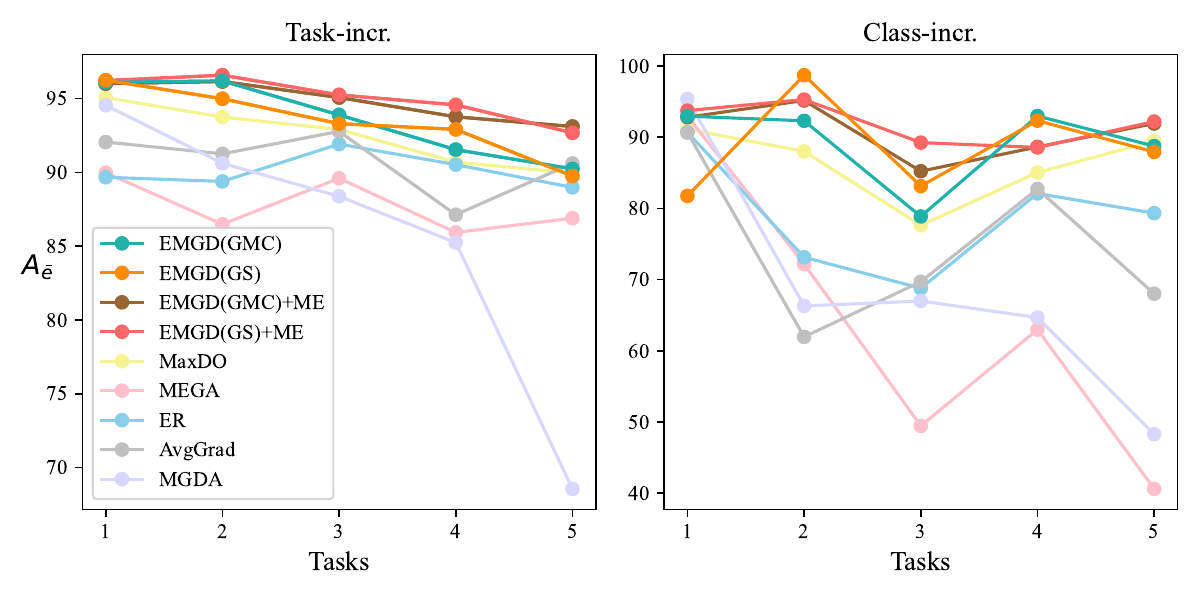}}
\subfigure[PS-CIFAR]{\centering
  \includegraphics[width=\linewidth]{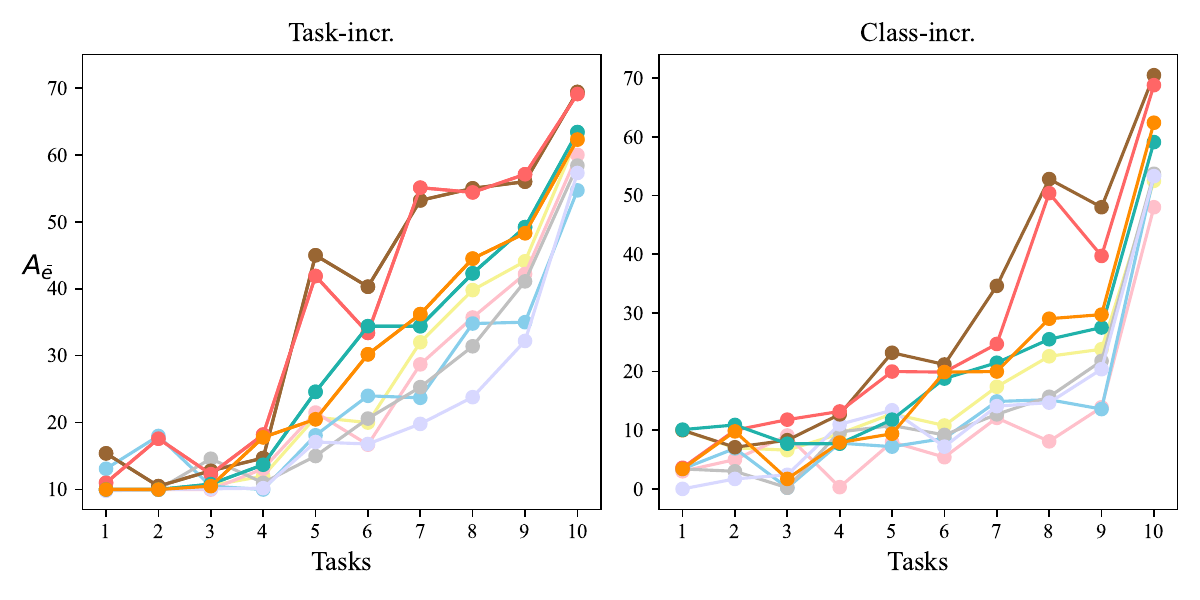}}
\subfigure[PS-ImageNet-TINY]{\centering
  \includegraphics[width=\linewidth]{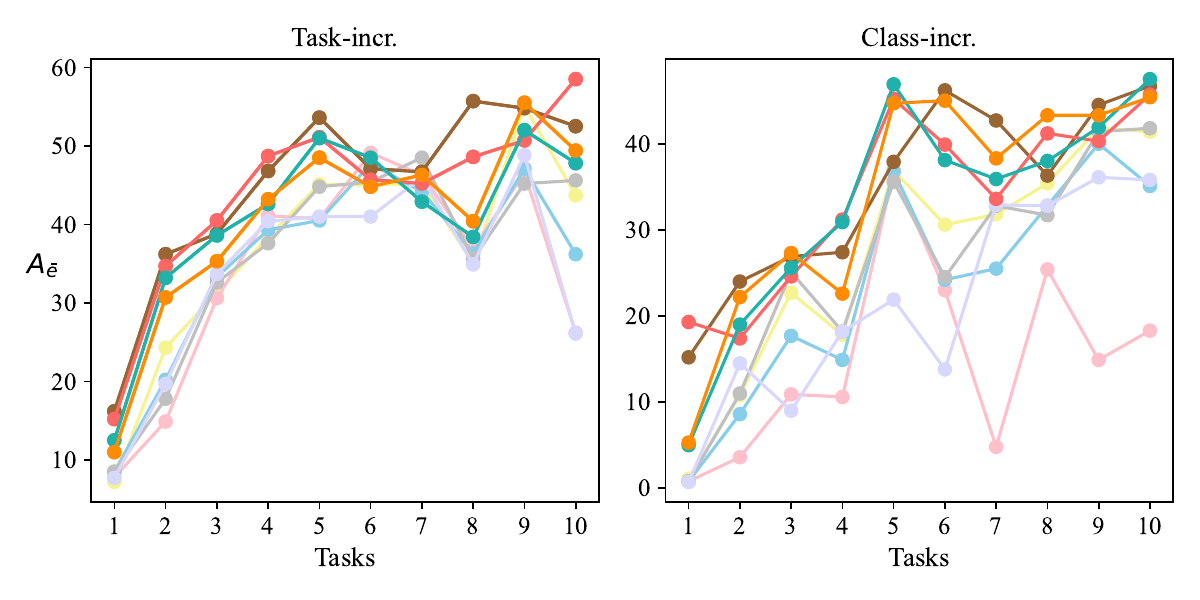}}
\caption{Learning process comparisons on three datasets.}
\label{fig:7}
\end{figure}

\subsection{Main Results}

We compare EMGD with MTL methods including MGDA~\cite{sener2018multi}, GradNorm~\cite{chen2018gradnorm}, DWA~\cite{liu2019end}, GradDrop~\cite{chen2020just}, PCGrad~\cite{yu2020gradient} and RLW~\cite{lin2021closer}.
Though the MTL methods are not designed for PCL, we treat any time as an MTL subunit and the connection of all these subunits can be used to train PCL.
We also compare our method with AvgGrad as illustrated in Sec.~\ref{sec:dmtl}. 
We also compare EMGD with SCL methods including AGEM~\cite{AGEM}, GMED~\cite{GMED}, MEGA~\cite{guo2019learning} and MDMTR~\cite{lyu2021multi}.
To adapt to the SCL methods, we merge batches from every in-training task to mimic a naive sequence learning that only a batch from the current task and a batch from the memory buffer.
For our proposed method, we use EMGD(GMC) and EMGD(GS) to represent EMGD using gradient magnitude change or gradient similarity to determine elastic factors.

As shown in Tables~\ref{tab:main_task} and \ref{tab:main_class}, we compare our method with MTL and SCL methods on three datasets in the task- and class-incremental PCL scenarios, respectively.
We have several major observations.
First, the final accuracy $A_{\bar{e}}$ of EMGD is better than that of the compared methods on three datasets, which shows the superiority of EMGD in solving task conflicts and catastrophic forgetting.
Second, with the proposed gradient-guided Memory Editing (ME), the performance of EMGD can be further improved, which means that with the guidance of a good gradient, the memory may reduce its negative influence on future training conflicts.
Third, similar to SCL, task-incremental PCL has less forgetting than class-incremental PCL.
We think that this is because task-incremental has specific task IDs for training, which may reduce the knowledge negative transfer for old tasks.
However, for class-incremental PCL, we find that forgetting may be more important than training conflict.  
The third finding is AvgGrad is good enough compared to other methods in the PCL setting.
This means the compared MTL and SCL methods focus on only studying one of the balanced training and less forgetting, which is not appropriate for the PCL setting.
Moreover, MGDA has poor performance in task-incremental PCL but good in class-incremental PCL, which also shows that task-incremental PCL focuses more on training conflict and class-incremental PCL should first solve forgetting.
Our EMGD considers the diverse training process in PCL and designs to obtain a Pareto descent direction, which balances the training between new task learning and old task keeping.
Memory editing helps the training towards better balance in future training.
Finally, we compare two metrics, GMC and GS, and observe that in task-incremental PCL, GS outperforms GMC. However, in class-incremental PCL, GMC performs better than GS. This indicates that in task-incremental PCL, the availability of task IDs allows the model to focus more on individual tasks. On the other hand, in class-incremental PCL, considering task correlations becomes more crucial for improving performance.

\subsection{Reherasal analysis of EMGD}

We evaluate PCL with different sizes of rehearsal coreset on PS-EMNIST and PS-CIFAR-100 datasets.
The results are shown in Fig.~\ref{fig:var_mem_2}.
We have the following observations.
First, with the size growth of the memory buffer, EMGD has a better $A_{\bar{e}}$, which means the retraining on rehearsal data streams can clearly reduce the catastrophic forgetting in PCL.
As the memory size increases, the rate of performance improvement in PCL slows down, and in some cases, it may even lead to a slight decline in performance.
Larger memory sizes have a more pronounced impact on performance enhancement with GS, while GMC demonstrates better performance at smaller memory sizes.

\subsection{{Visualization of edited gradient}}


In Fig.~\ref{fig:vis}, we conducted visualization on gradient editing (normalized gradients in Eq.~\eqref{eq:me}) using two datasets: PS-EMNIST and PS-CIFAR-100. 
Brighter pixels mean more edits.
We observe a phenomenon where the ease of editing gradients was influenced by the information density of the data.
On the PS-EMNIST dataset, we find that the gradients were more easily edited in regions with lower information density. This implies that the areas of the background are more susceptible to being edited. 
On the other hand, in the PS-CIFAR-100 dataset, which contains complex images from various categories, we noticed that gradients associated with higher information density pixels are more prone to be edited. These pixels contained more intricate features, textures, and patterns, making them more likely to be altered. 
These experimental results suggest that on simple datasets, where the task information is more explicit and the density of information is higher, rehearsal performs better. 
However, on more complex datasets, where there is greater uncertainty in the data, it is necessary to selectively edit the key elements of the original data to enhance its compatibility with future tasks. This targeted editing helps to improve the data's ability to support subsequent tasks by reducing uncertainty and enhancing relevant information.








\begin{figure}[t]
\centering
\centering
  \includegraphics[width=\linewidth]{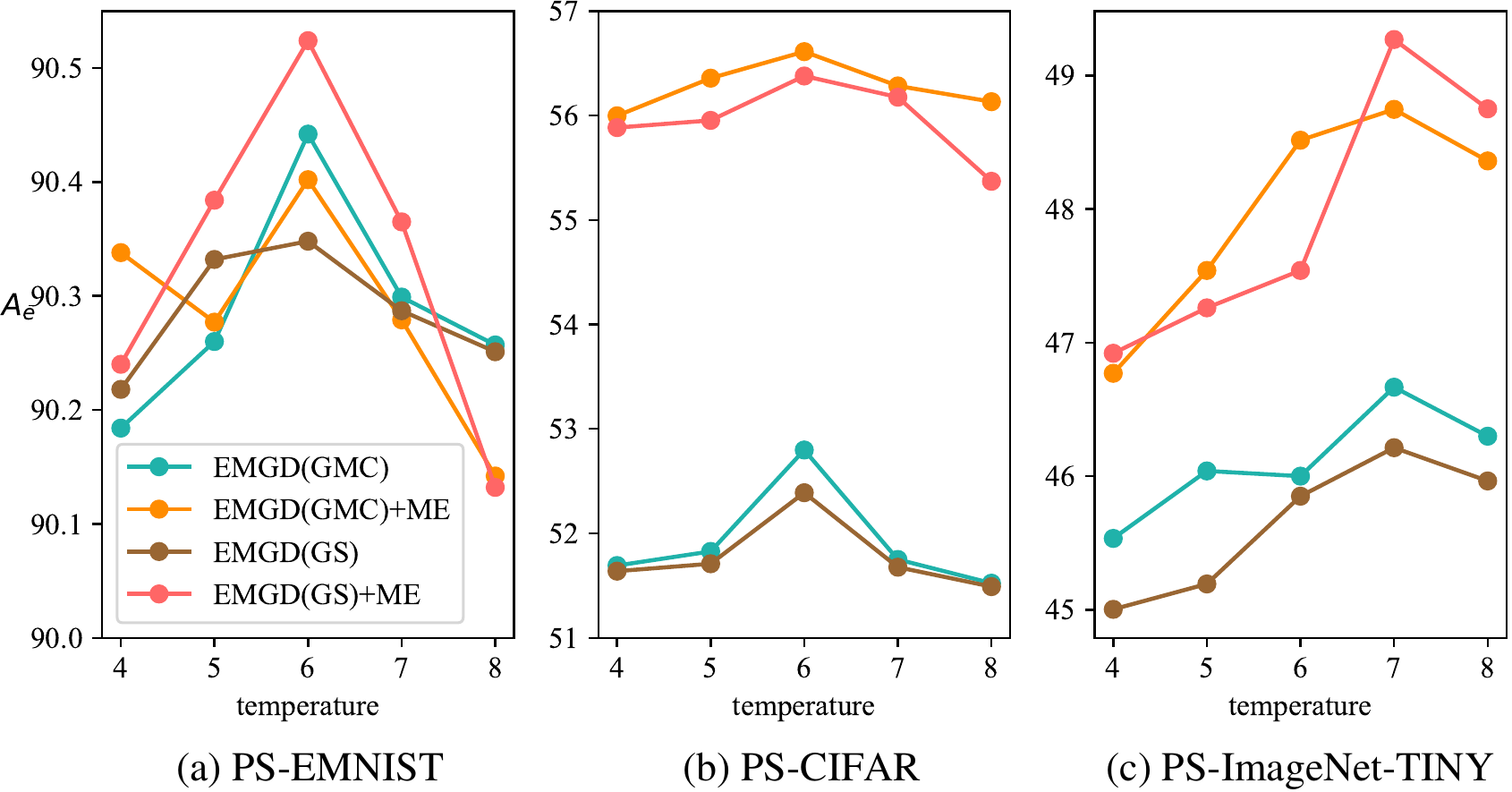}
\caption{Softmax temperature selection for three datasets.}
\label{fig:8}
\end{figure} 

\begin{table}[t]
  \centering
  \caption{
    Multi-epoch experiments on PS-CIFAR-100.
  }
  \resizebox{\linewidth}{!}{
    \begin{tabular}{l|cc|cc}
      \toprule
      \multirow{2}{*}{\textbf{Method}} &  \multicolumn{2}{c|}{\textbf{Task increment}} &  \multicolumn{2}{c}{\textbf{Class increment}} \\
       & $A_\text{first}$ (\%) & $F_\text{final}$ (\%)  & $A_\text{first}$ (\%) & $F_\text{final}$ (\%)  \\ 
      \midrule
      AvgGrad 		& {59.600} $\pm$ 0.437 & {-5.440} $\pm$ 0.271 & {19.390} $\pm$ 0.366	& {-17.990} $\pm$ 0.650  \\
      MGDA 		& {58.780} $\pm$ 6.844 & {-2.770} $\pm$ 6.250 & {21.057} $\pm$ 1.269	& {-14.886} $\pm$ 0.810  \\
      PCGrad 		& {56.720} $\pm$ 1.712 & {-6.210} $\pm$ 0.962 & {17.610} $\pm$ 0.105	& {-24.000} $\pm$ 1.422  \\

      AGEM 		& {49.949} $\pm$ 1.533 & {-14.084} $\pm$ 1.779 & {7.408} $\pm$ 0.411	& {-35.003} $\pm$ 0.896  \\
      ER 		& {59.322} $\pm$ 0.349 & {-6.774} $\pm$ 0.619 & {16.030} $\pm$ 0.787	& {-21.300} $\pm$ 0.058  \\
      MEGA 		& {60.710} $\pm$ 0.386 & {-2.779} $\pm$ 0.518 & {22.710} $\pm$ 0.317	& {-11.470} $\pm$ 0.731  \\
      MaxDO 		& {60.400} $\pm$ 1.862 & {-0.997} $\pm$ 2.326 & {22.940} $\pm$ 0.727	& {-19.020} $\pm$ 2.976  \\
      \midrule
      EMGD(GMC) 		& {63.754} $\pm$ 1.080 & {-3.488} $\pm$ 0.862 & {26.984} $\pm$ 0.328	& {-11.944} $\pm$ 1.238  \\
      EMGD(GMC)+ME 		& {65.222} $\pm$ 0.420 & {-2.586} $\pm$ 0.403 & \textbf{30.657} $\pm$ 0.263	& \textbf{-3.483} $\pm$ 0.707  \\
      EMGD(GS) 		& {64.462} $\pm$ 0.524 & {-2.255} $\pm$ 0.734 & {25.885} $\pm$ 0.069	& {-8.741} $\pm$ 0.660  \\
      EMGD(GS)+ME 		& \textbf{67.260} $\pm$ 0.926 & \textbf{-1.470} $\pm$ 0.143 & {29.445} $\pm$ 0.281	& {-2.607} $\pm$ 0.973  \\
      \bottomrule
    \end{tabular}}
  \label{tab:multiepoch}
\end{table}

\begin{figure*}[t]
\centering
\subfigure[TASK 1 - early (EMGD) ]{\centering
  \includegraphics[width=0.32\textwidth]{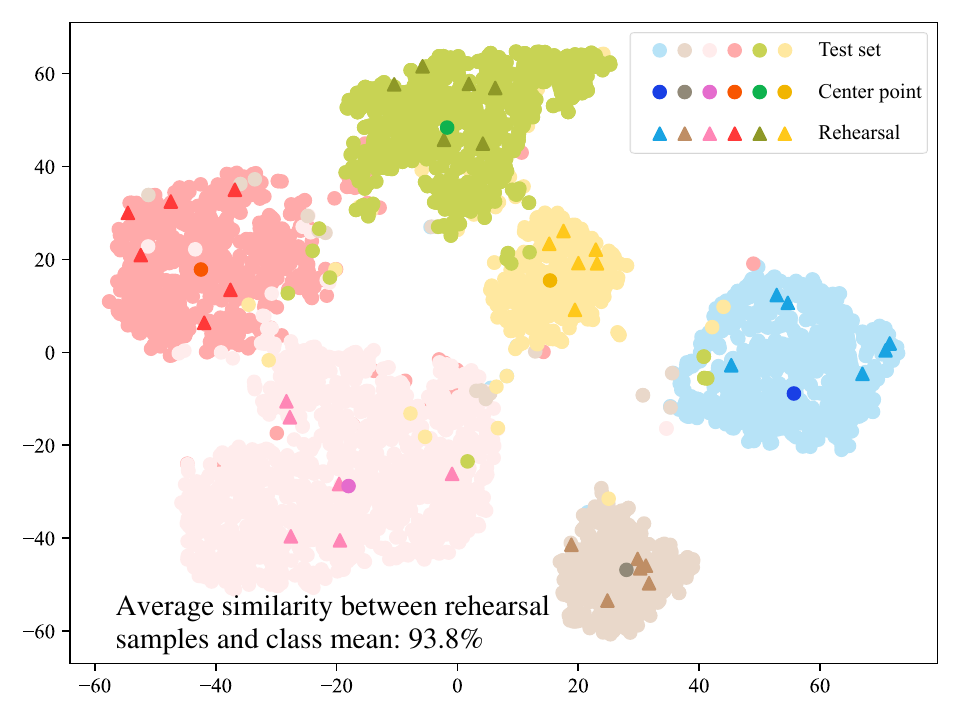}}
\subfigure[Task 1 - final (ER)]{\centering
  \includegraphics[width=0.32\textwidth]{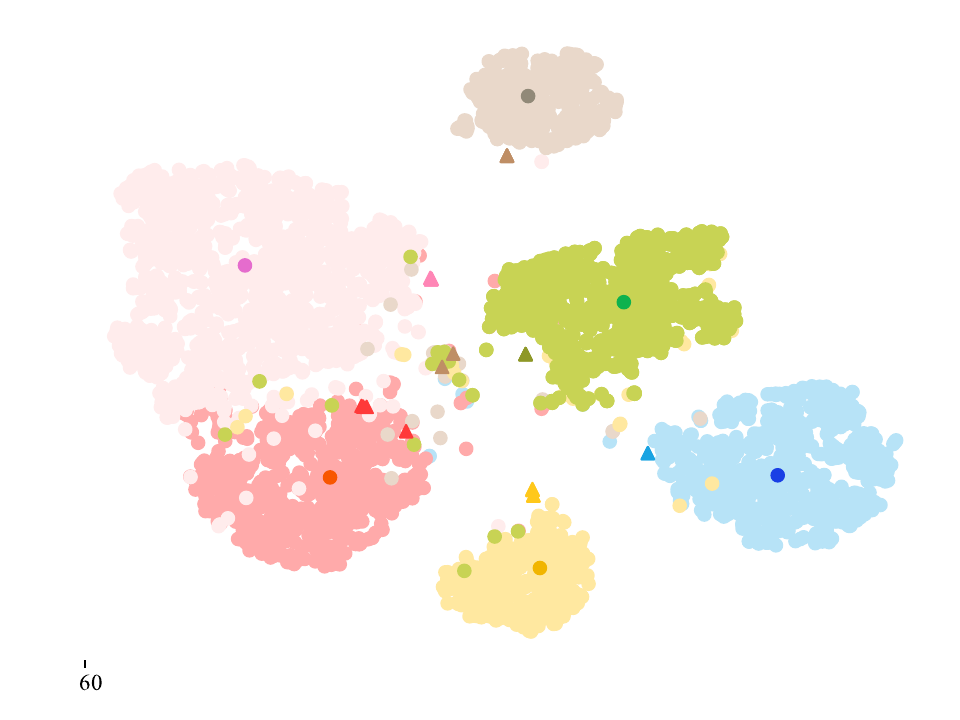}}
\subfigure[Task 1 - final (EMGD+ME)]{\centering
  \includegraphics[width=0.318\textwidth]{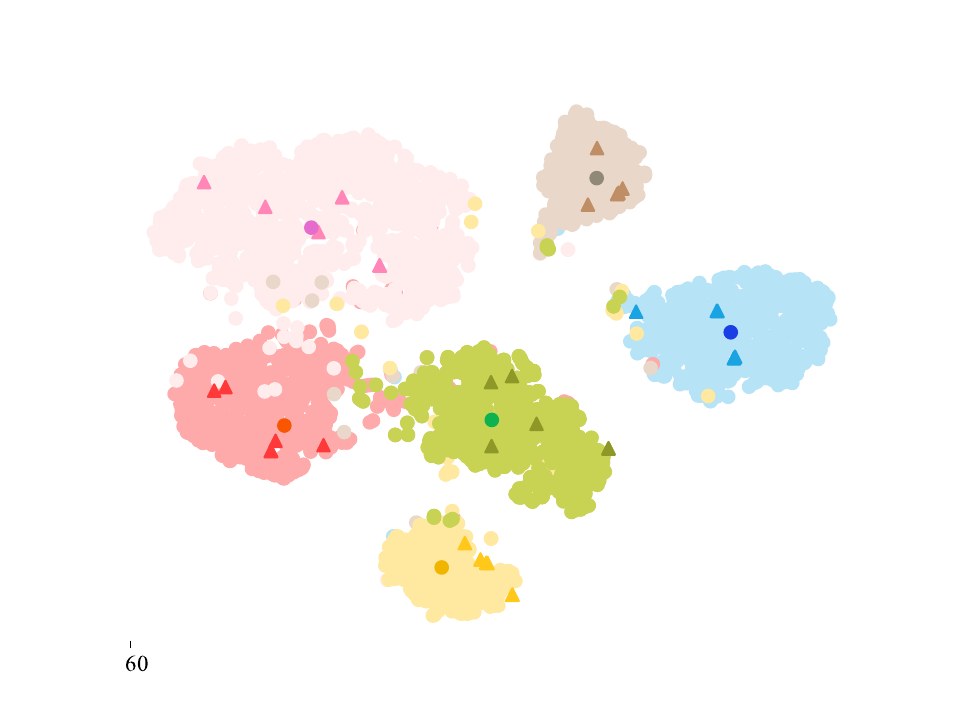}}
\caption{t-SNE visualization on PS-EMNIST. We omit the ``Task 1 - early (ER)'' cause it is quite similar to (a) TASK 1 - early (EMGD) at the early of PCL.}
\label{fig:9}
\end{figure*}

\begin{table}[t]
  \centering
  \caption{
    Blurring PCL Evaluation on CIFAR-100.
  }
  \resizebox{\linewidth}{!}{
    \begin{tabular}{l|cc|cc}
      \toprule
      \multirow{2}{*}{\textbf{Method}} &  \multicolumn{2}{c|}{\textbf{Task increment}} &  \multicolumn{2}{c}{\textbf{Class increment}} \\
       & $A_\text{first}$ (\%) & $F_\text{final}$ (\%)  & $A_\text{first}$ (\%) & $F_\text{final}$ (\%)  \\ 
      \midrule
      AvgGrad 		& {31.365} $\pm$ 0.767 & {18.070} $\pm$ 0.532 & {8.539} $\pm$ 0.253	& {-1.156} $\pm$ 0.169  \\
      MGDA 		& {31.492} $\pm$ 0.095 & {18.088} $\pm$ 0.337 & {11.220} $\pm$ 0.684	& {1.827} $\pm$ 0.749  \\
      PCGrad 		& {31.732} $\pm$ 1.636 & {18.276} $\pm$ 1.436 & {8.874} $\pm$ 0.510	& {-1.042} $\pm$ 0.490  \\

      AGEM 		& {18.502} $\pm$ 0.868 & {7.584} $\pm$ 0.244 & {9.477} $\pm$ 0.526	& {4.431} $\pm$ 0.293  \\
      ER 		& {31.044} $\pm$ 0.625 & {17.744} $\pm$ 0.056 & {8.142} $\pm$ 0.263	& {-1.511} $\pm$ 0.040  \\
      MEGA 		& {30.809} $\pm$ 0.336 & {18.074} $\pm$ 0.278 & {11.567} $\pm$ 0.414	& {4.051} $\pm$ 0.665  \\
      MaxDO 		& {32.391} $\pm$ 1.519 & {18.382} $\pm$ 1.305 & {10.912} $\pm$ 0.321	& {-0.271} $\pm$ 0.747  \\
      \midrule
      EMGD(GMC) 		& {34.860} $\pm$ 0.975 & {20.244} $\pm$ 0.881 & {12.473} $\pm$ 0.115	& {1.785} $\pm$ 0.272  \\
      EMGD(GMC)+ME 		& \textbf{45.139} $\pm$ 1.032 & \textbf{24.593} $\pm$ 1.428 & {18.227} $\pm$ 0.466	& {1.887} $\pm$ 0.264  \\
      EMGD(GS) 		& {34.811} $\pm$ 1.685 & {20.212} $\pm$ 1.546 & {12.861} $\pm$ 0.904	& {2.089} $\pm$ 0.930  \\
      EMGD(GS)+ME 		& {44.719} $\pm$ 0.840 & {24.321} $\pm$ 0.509 & \textbf{18.912} $\pm$ 0.793	& \textbf{2.742} $\pm$ 0.581  \\
      \bottomrule
    \end{tabular}}
  \label{tab:blur}
\end{table}

\subsection{{PCL learning process}}

As shown in Fig.~\ref{fig:7}, we evaluate the performance on seen tasks along the whole PCL training process.
The point $i$ in Fig.~\ref{fig:7} means that the task $i$ just finished its training, and the corresponding value denotes the average testing accuracy on all seen tasks (\ie, tasks in training and finished). 
Our experiment effectively enhanced overall performance in three datasets and two PCL settings (task-incremental and class-incremental PCL). 
Our method consistently outperforms baseline models across datasets, validating its robustness and generalizability. It demonstrated remarkable consistency in performance, ensuring effectiveness throughout the entire PCL training process. 
These findings show that our EMGD method contributes to PCL problem, providing a reliable solution for optimizing performance.
Moreover, we compare two kinds of computing of elastic factors.
We find that GMC has better performance on the early training tasks while GS has better performance on the last training period.
We think this is because, in the early training period, the number of seen tasks is small, and the training of each task considers more about itself. 
While in the last training period, more tasks need to be considered, the gradient similarity is important to adjust the conflict.

\subsection{{Hyperparameter selection}}

In Fig.~\ref{fig:8}, we show the major hyperparameter selection for Softmax temperature in GMC (Eq.~\eqref{eq:sigma}) and GS (Eq.~\eqref{eq:sigma_2}). 
The results show that the trends of different strategy are similar, and
the two strategies have similar sensitivity to temperature.
Therefore, it is feasible to select the same hyperparameter for the two strategies by tuning once.

\subsection{{Visualization of feature space}}


As depicted in Fig.~\ref{fig:9}, we employed the t-SNE~\cite{van2008visualizing} tool to visualize the feature space and demonstrate the effectiveness of our proposed memory editing technique. The visualizations clearly indicate that the proposed memory editing method successfully maintains representation and uniformity. In contrast, due to semantic drift, the traditional ER method struggles to maintain uniformity during the PCL process. Furthermore, we computed the average cosine similarity between rehearsal samples and class means. After task 1 is fully trained, there is a considerable similarity between randomly selected samples and class means. However, as all tasks complete their training, the similarity in ER decreases significantly due to semantic drift, where some classes have their rehearsal samples out of distributions. In contrast, our memory editing method effectively preserves the similarity, demonstrating that our rehearsal approach mitigates forgetting to a greater extent.


\subsection{{Multi-epoch comparisons}}

In major experiments, we evaluate in the 1-epoch online setting for each task.
In Table \ref{tab:multiepoch}, we evaluate in 20-epoch for each task without changing the training order.
The results show that the 20-epoch setting provides better final accuracy than the single-epoch setting because each task can be trained more sufficiently toward convergence.
Moreover, we find that forgetting becomes harder than before because more new task training is implemented after the old tasks are finished.
Compared to the other methods, our 20-epoch performances are also better than them and even further expand the advantages.
The results show that our methods are also effective in off-line or multi-epoch scenarios.


\subsection{{EMGD in Blurring PCL}}

Blurring CL~\cite{bang2021rainbow} is different from traditional CL setting, where the label spaces are disjoint for any two tasks.
Blurring CL has overlap label spaces to meet the real-world needs.
We also implement blurring PCL on CIFAR-100.
Specifically, we split the 100 classes of CIFAR-100 into 9 overlap tasks, each of which has 20 classes.
The label space has 50\% overlap between any two neighbouring tasks.
The results are shown in Table~\ref{tab:blur}, which shows that our EMGD is also effective on blurring PCL.
For the two metrics of evaluation elastic factors, we find that GMC is more effective on task-incremental blurring PCL setting while GS is more effective on class-incremental blurring PCL.
This is because that no taks-id in class-incremental setting, the gradient similarity is useful to adjust the influence from one task to another task.  
The results also show that the proposed memory editing is useful for rehearsal-based blurring PCL.

\begin{table*}[t]
	\centering
	\caption{
	  {Comparisons on SCL dataset Split-CIFAR-10, Split-CIFAR-100 and Split-Tiny-ImageNet.}
	}
	\resizebox{\linewidth}{!}{
	  \begin{tabular}{l|rr|rr|rr}
		\toprule
		\multirow{2}{*}{\textbf{Method}} &  \multicolumn{2}{c|}{\textbf{Split-CIFAR-10}} &  \multicolumn{2}{c|}{\textbf{Split-CIFAR-100}} &  \multicolumn{2}{c}{\textbf{Split-Tiny-Imagenet}}\\
		 & \multicolumn{1}{c}{$A_{\bar{e}}$ (\%)} & \multicolumn{1}{c|}{$F_{\bar{e}}$ (\%)}  & \multicolumn{1}{c}{$A_{\bar{e}}$ (\%)} & \multicolumn{1}{c|}{$F_{\bar{e}}$ (\%)}& \multicolumn{1}{c}{$A_{\bar{e}}$ (\%)} & \multicolumn{1}{c}{$F_{\bar{e}}$ (\%)} \\  
		\midrule
		iCarL~\cite{rebuffi2017icarl}&$74.917\pm1.579$&$0.013\pm0.938$&$31.960\pm0.715$&$0.740\pm0.639$&\multicolumn{1}{c}{-}&\multicolumn{1}{c}{-}\\
		GEM~\cite{lopez2017gradient}	&$73.027\pm3.571$&$0.167\pm4.344$&$39.380\pm1.914$&$4.407\pm1.434$&\multicolumn{1}{c}{-}&\multicolumn{1}{c}{-}\\
		AGEM~\cite{AGEM}	&$71.700\pm2.644$&$-3.533\pm2.796$&$32.607\pm2.790$&$-2.960\pm2.779$&$20.733\pm0.682$&$-8.853\pm0.439$\\
		GDUMB~\cite{GDUMB}	       & $50.937\pm1.254$ & \multicolumn{1}{c|}{-} & $17.720\pm0.732$ & \multicolumn{1}{c|}{-} &$6.878\pm0.955$& \multicolumn{1}{c}{-}  \\ 
		GSS~\cite{GSS}          &$68.663\pm2.761$&$-5.603\pm3.141$&$32.757\pm0.526$&$-8.753\pm1.673$&$27.000\pm1.578$&$-9.419\pm0.943$\\
		MIR~\cite{MIR}   &$74.900\pm3.034$&$-1.163\pm1.480$&$43.140\pm3.840$&$\mb{6.833}\pm3.456$&$27.961\pm0.310$&$-1.983\pm0.192$\\
		GMED~\cite{GMED}  &$75.983\pm1.121$&${0.853}\pm1.351$&$44.697\pm4.934$&$2.930\pm4.574$&$27.911\pm1.015$&$-0.936\pm2.124$\\
		MaxDO~\cite{lyu2023measuring} &$74.324\pm2.402$&$-2.835\pm2.588$&$44.809\pm1.910$&$4.332\pm1.004$&$28.380\pm0.839$&$-0.610\pm0.536$ \\ 
		\midrule
		EMGD(GMC) 		&${76.983}\pm0.268$&$-0.970\pm1.967$&${47.037}\pm0.397$&$3.920\pm1.204$&${31.067}\pm0.707$&${0.508}\pm0.661$\\
        EMGD(GMC)+ME   &${77.356}\pm0.789$&$0.234\pm1.198$&$\mb{49.500}\pm0.876$&$5.123\pm0.456$&$\mb{33.587}\pm0.319$&$\mb{0.745}\pm0.862$\\
        EMGD(GS) 		       &${76.438}\pm0.597$&$-0.675\pm1.256$&${47.932}\pm0.104$&$3.879\pm0.619$&${31.403}\pm0.758$&${0.531}\pm0.287$\\
        EMGD(GS)+ME 	   &$\mb{77.695}\pm0.374$&$\mb{0.922}\pm1.569$&${49.218}\pm0.746$&$4.385\pm0.654$&${33.421}\pm0.879$&${0.593}\pm0.127$\\
		\bottomrule
	  \end{tabular}}
	\label{tab:scl}
  \end{table*}

\begin{table}[t]
  \centering
  \caption{
    MTL Comparisons on Split-CIFAR-10, Split-CIFAR-100, and Split-Tiny-ImageNet.
  }
  \resizebox{\linewidth}{!}{
    \begin{tabular}{l|ccc}
      \toprule
      \textbf{Method} &  \textbf{Split-CIFAR10} &  \textbf{Split-CIFAR100} & \textbf{Split-Tiny}\\
      \midrule
      AvgGrad	                     & 76.127 $\pm$ 0.502 & 23.303 $\pm$ 1.626 & 37.840 $\pm$ 1.789\\
      MGDA~\cite{sener2018multi}     & 73.213 $\pm$ 1.071 & 23.320 $\pm$ 2.015 & 35.923 $\pm$ 1.007\\ 
      GradNorm~\cite{chen2018gradnorm}     & 76.127 $\pm$ 0.799 & 21.897 $\pm$ 4.882 & 36.890 $\pm$ 0.178\\
      DWA~\cite{liu2019end}          & 76.217 $\pm$ 0.515 & 23.417 $\pm$ 1.568 & 36.420 $\pm$ 2.841\\
      GradDrop~\cite{chen2020just} & 75.810 $\pm$ 0.470 & 23.403 $\pm$ 1.460 & 38.110 $\pm$ 0.760\\
      PCGrad~\cite{yu2020gradient}    & 76.237 $\pm$ 0.592 & 22.910 $\pm$ 1.762 & 37.743 $\pm$ 1.497\\
      RLW~\cite{lin2021closer} 		      & 76.100 $\pm$ 0.155 & 23.243 $\pm$ 1.589 & 39.253 $\pm$ 0.993\\
    MaxDO~\cite{lyu2023measuring} &76.299 $\pm$ 0.443 & 23.546 $\pm$ 1.671 & 39.297 $\pm$ 0.764 \\ 
      \midrule
      EMGD(GMC) 		 & {76.373} $\pm$ 1.472 & {24.147} $\pm$ 2.110 & {39.467} $\pm$ 0.722\\
      EMGD(GS) 		 & \textbf{76.986} $\pm$ 1.304 & \textbf{24.780} $\pm$ 2.456 & \textbf{39.842} $\pm$ 0.635\\
      \bottomrule
    \end{tabular}}
  \label{tab:mtl}
\end{table}

\subsection{EMGD in SCL}

Although EMGD is designed for PCL where multiple tasks are trained together.
In this subsection, we also show that EMGD can be used in rehearsal-based SCL, where the rehearsal plays the role of finished task.
That is, we will always have two parallel tasks from the second task.
To verify this, in the SCL setting, we compare EMGD with 7 rehearsal-based methods (iCarL~\cite{rebuffi2017icarl}, GDUMB~\cite{GDUMB}, GEM~\cite{lopez2017gradient}, AGEM~\cite{AGEM}, GSS~\cite{GSS}, MIR~\cite{MIR} and GMED~\cite{GMED}). 
We evaluate on two popular public SCL datasets, \ie, Split-CIFAR-10~\cite{zenke2017continual} and Split-CIFAR-100~\cite{zenke2017continual}.
We store 300 and 3,000 in total for Split-CIFAR-10 and Split-CIFAR-100, respectively.
Each data point in the online fashion will pass only once.
{We also use $A_{\bar{e}}$ and $F_{\bar{e}}$ to evaluate the capacity of models.}
We show the results in Table~\ref{tab:scl}.
{EMGD outperforms most of the compared methods, especially in $A_{\bar{e}}$}.
This means that EMGD can well balance the memory task and the new task in SCL process.
Another observation is that online fashion may lead the training insufficient, and continual training on memory may further improve the performance then drop after convergence.

\subsection{EMGD in MTL}

In this subsection, we show that the proposed EMGD can also be applied to common Multi-Task Learning (MTL).
We compare EMGD with SOTA MTL methods including MGDA~\cite{sener2018multi}, GradNorm~\cite{chen2018gradnorm}, DWA~\cite{liu2019end}, GradDrop~\cite{chen2020just}, PCGrad~\cite{yu2020gradient}, IMTL~\cite{liu2021towards} and RLW~\cite{lin2021closer} in MTL setting.
In our experiments, we also use the Split-CIFAR-10, Split-CIFAR-100 and Split-Tiny-Imagenet. 
The results are shown in Table~\ref{tab:mtl}.
It is clear that even though MTL suffers less from the diverse training process challenges, gradients have small differences at each training step.
EMGD captures the small gradient differences and adjusts the Pareto descent direction and achieves better performance than other methods.

\section{Conclusion and Future Work}

In this paper, we compared traditional Multi-Task Learning (MTL) and Serial Continual Learning (SCL) that train a single model for multi-task on multi-source data. 
MTL lacks the ability to adaption to new knowledge while SCL can not respond to new tasks immediately.
In this paper, we proposed a novel and practical continual learning problem, called Parallel Continual Learning (PCL), in which the new task can be trained with previous tasks parallelly without pending.
To optimize the model among dynamic multiple tasks, we treat PCL as a dynamic multi-objective optimization problem.
Based on the well-proven steepest descent method, we propose to set appropriate elastic factors to constrain the training of each task.
We named our method Elastic Multi-Gradient Descent (EMGD).
To reduce the task conflict inside memory, we also proposed a gradient-guided memory editing method. 
The experimental results on three image recognition clearly show the effectiveness of the proposed benchmark.

\lyu{
At any time, the multiple tasks in PCL are trained in parallel.
Although our proposed method achieved clear performance improvement compared to other methods, we assume the training is only one one client or machine.
In our view, PCL has latent relationship with parallel computing and distributed learning. 
It is foreseeable to further accelerate PCL training via combining these techniques to meet the requirements from real-world applications.}


\nocite{langley00}


\bibliography{ref}
\bibliographystyle{abbrv}








\end{document}